\documentclass[10pt,twocolumn,letterpaper]{article}

\usepackage{iccv}
\usepackage{times}
\usepackage{epsfig}
\usepackage{graphicx}
\usepackage{amsmath}
\usepackage{amssymb}
\usepackage{booktabs}
\usepackage[dvipsnames,table,xcdraw]{xcolor}
\usepackage[utf8]{inputenc} %
\usepackage[T1]{fontenc}    %
\usepackage{url}            %
\usepackage{amsfonts}       %
\usepackage{nicefrac}       %
\usepackage{microtype}      %
\usepackage{multirow}
\usepackage{bbm}
\usepackage{rotating}
\usepackage[english]{babel}
\usepackage{pifont}%
\usepackage[dvipsnames]{xcolor}
\usepackage[caption=false]{subfig}

\usepackage[pagebackref=true,breaklinks=true,letterpaper=true,colorlinks,bookmarks=false]{hyperref}

\newcommand{\mysection}[1]{\vspace{3pt}\noindent\textbf{#1.}}

\newcommand{\sota}{state-of-the-art\xspace}

\usepackage{listings}
\usepackage{color}
\definecolor{codegreen}{rgb}{0,0.6,0}
\definecolor{codegray}{rgb}{0.5,0.5,0.5}
\definecolor{codepurple}{rgb}{0.58,0,0.82}
\definecolor{backcolour}{rgb}{1,1,1}
 
\lstdefinestyle{mystyle}{
    backgroundcolor=\color{backcolour},   
    commentstyle=\color{codegreen},
    keywordstyle=\color{magenta},
    numberstyle=\tiny\color{codegray},
    stringstyle=\color{codepurple},
    basicstyle=\footnotesize,
    breakatwhitespace=false,         
    breaklines=true,                 
    captionpos=b,                    
    keepspaces=true,                 
    numbers=left,                    
    numbersep=5pt,                  
    showspaces=false,                
    showstringspaces=false,
    showtabs=false,                  
    tabsize=2
}
 
\iccvfinalcopy %

\def\iccvPaperID{5299} %

\begin{document}

\newcommand{\authorsep}{\hspace{8pt}}

\title{EgoLoc: Revisiting 3D Object Localization from Egocentric Videos \\ with Visual Queries}

\author{Jinjie Mai$^1$ \authorsep Abdullah Hamdi$^{2,1}$ \authorsep Silvio Giancola$^1$ \authorsep Chen Zhao$^1$  \authorsep Bernard Ghanem$^1$\\
\normalsize$^1$King Abdullah University of Science and Technology (KAUST) \hspace{5pt} \normalsize$^2$Visual Geometry Group, University of Oxford
\\
{\tt\small \{jinjie.mai,silvio.giancola,chen.zhao,bernard.ghanem\}@kaust.edu.sa}\\
\tt\small{abdullah.hamdi@eng.ox.ac.uk}
}

\maketitle

\begin{abstract}
    With the recent advances in video and 3D understanding, novel 4D spatio-temporal methods fusing both concepts have emerged.
    Towards this direction, the Ego4D Episodic Memory Benchmark proposed a task for Visual Queries with 3D Localization (VQ3D).
    Given an egocentric video clip and an image crop depicting a \textit{query object}, the goal is to localize the 3D position of the center of that \textit{query object} with respect to the camera pose of a \textit{query frame}.
    Current methods tackle the problem of VQ3D by unprojecting the 2D localization results of the sibling task Visual Queries with 2D Localization (VQ2D) into 3D predictions. 
    Yet, we point out that the low number of camera poses caused by camera re-localization from previous VQ3D methods severally hinders their overall success rate.
    In this work, we formalize a pipeline (we dub \textbf{EgoLoc}) that better entangles 3D multiview geometry with 2D object retrieval from egocentric videos. 
    Our approach involves estimating more robust camera poses and aggregating multi-view 3D displacements by leveraging the 2D detection confidence, which enhances the success rate of object queries and leads to a significant improvement in the VQ3D baseline performance. 
    Specifically, our approach achieves an overall success rate of up to 87.12\%, which sets a new state-of-the-art result in the VQ3D task\footnote{\url{https://eval.ai/web/challenges/challenge-page/1646/leaderboard/3947}}. 
    We provide a comprehensive empirical analysis of the VQ3D task and existing solutions, and highlight the remaining challenges in VQ3D. The code is available at \url{https://github.com/Wayne-Mai/EgoLoc}.
    
\end{abstract}

\begin{figure}[t]
    \centering
    \includegraphics[width=\linewidth] {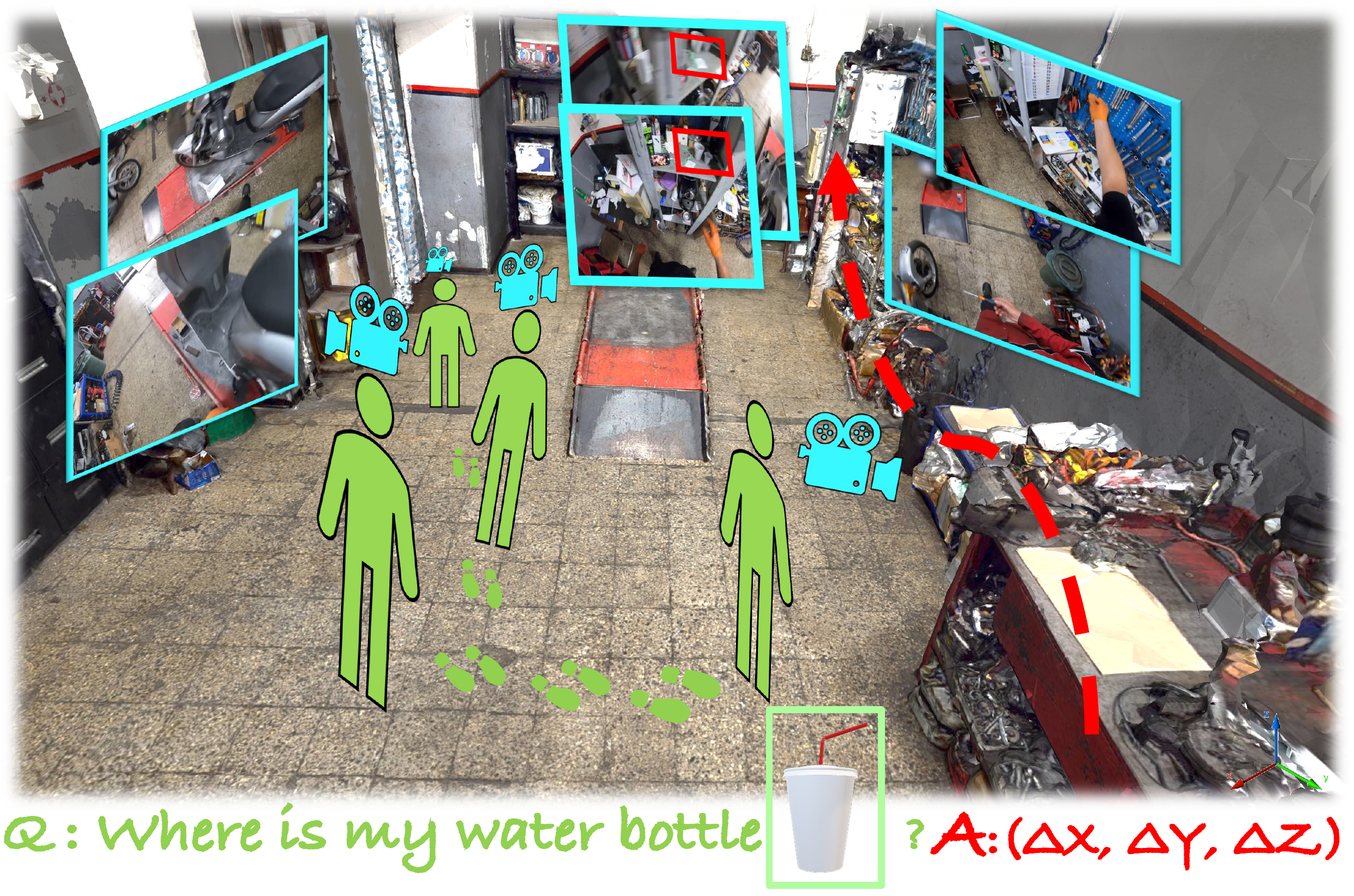}
    \caption{
        \textbf{Visual Query with 3D Localization Task in Egocentric Videos}. 
        Given an egocentric video clip and an image crop depicting a \textit{query object}, the goal is to localize the last time a \textit{query object} was seen in the video and return the 3D displacement vector from the camera center of the \textit{query frame} to the center of the object in 3D. %
        }
    \label{fig:pull}
\end{figure}

\section{Introduction}
\label{sec:intro}

Time moves forward, from the past to the future, and we cannot turn it back. But our minds have a special ability to remember past events, almost as if we are traveling back in time. This ability is called \textit{Episodic Memory}, and it's unique to humans ~\cite{tulving2002episodic}. It's more than just remembering facts; it's about reliving past experiences, knowing when they happened, and understanding that they happened to us \cite{tulving2002episodic}.
In the pursuit of more human-like AI systems, infusing \textit{Episodic Memory} capabilities into our machines holds great promise, especially in assisting people to recall their past experiences.

Towards such efforts, the massive-scale dataset and benchmark suite Ego4D~\cite{grauman2022ego4d} introduced multiple tasks on \textit{Episodic Memory} from egocentric videos, with the scope of browsing and searching past human experiences.
Among those challenges, the task of Visual Queries (VQ) aimed at answering \textit{``Where was object $X$ last seen in the video?''}, with $X$ being a single image crop of an object, clearly visible and humanly identifiable.
In particular, Visual Queries with 3D Localization (VQ3D) focuses on retrieving the relative 3D localization of a \textit{query object} with respect to a current \textit{query frame}, as illustrated in Figure~\ref{fig:pull}.

The task of VQ3D arose from the natural progress in computer vision challenges, building on top of the latest development in image understanding~\cite{deng2009imagenet,zhao2019object}, video understanding~\cite{grauman2022ego4d,soldan2022mad}, and 3D geometric understanding~\cite{schoenberger2016sfm,mvtn,hamdi2023voint}.
Specifically, VQ3D requires a frame-wise understanding of an egocentric video to localize objects in 2D images, a special 2D localization of the object along the temporal dimension, coupled with a 3D scene understanding to unproject the 2D localization into a 3D environment.
Although most of the effort in Ego4D originates in the field of video understanding, little effort has been paid to improve the meaningful 3D knowledge needed by VQ3D methods.

Previous work~\cite{grauman2022ego4d} performs camera pose estimation by relocalizing real egocentric video frames to a Matterport scan, suffering from the \textit{ simulation-to-real} gap (difference in the domains and reference coordinates). IT also builds on 2D localization without proper 3D entanglement.
In this work, we attempt to bridge the gap between video and 3D scene understanding in VQ3D. 
In particular, we develop a pipeline that better entangles 3D multiview geometry with 2D object retrieval from egocentric videos. To fully understand the 3D scene, our proposed aggregation method predicts displacement from multiple views by leveraging the detection scores. This led to \sota results in the VQ3D task.     
We summarize our contributions as follows.

\begin{itemize}
    \item We formalize the pipeline for the task of Visual Queries with 3D Localization (VQ3D) from egocentric videos, with a thorough study of each module. We identify and solve the \textit{Simulation-2-Real} gap for camera pose estimation, and elevate the baseline performance from $8.71\%$~\cite{grauman2022ego4d} to $77.27\%$.
    
    \item We propose to aggregate multi-view 3D displacements by employing the 2D detection confidences to weight predictions and further enhance 3D localization. Our method (EgoLoc) achieves \textbf{87.12\%} in Overall Success Rate on the test set of the VQ3D task, significantly outperforming the baseline and setting new state-of-the-art results in VQ3D.

    \item We perform an extensive empirical analysis of different components and configurations in the VQ3D pipeline, which aims to benefit future research in the VQ3D direction.

\end{itemize}

\begin{figure*}[t]
    \centering
    \includegraphics[width=0.9\linewidth] {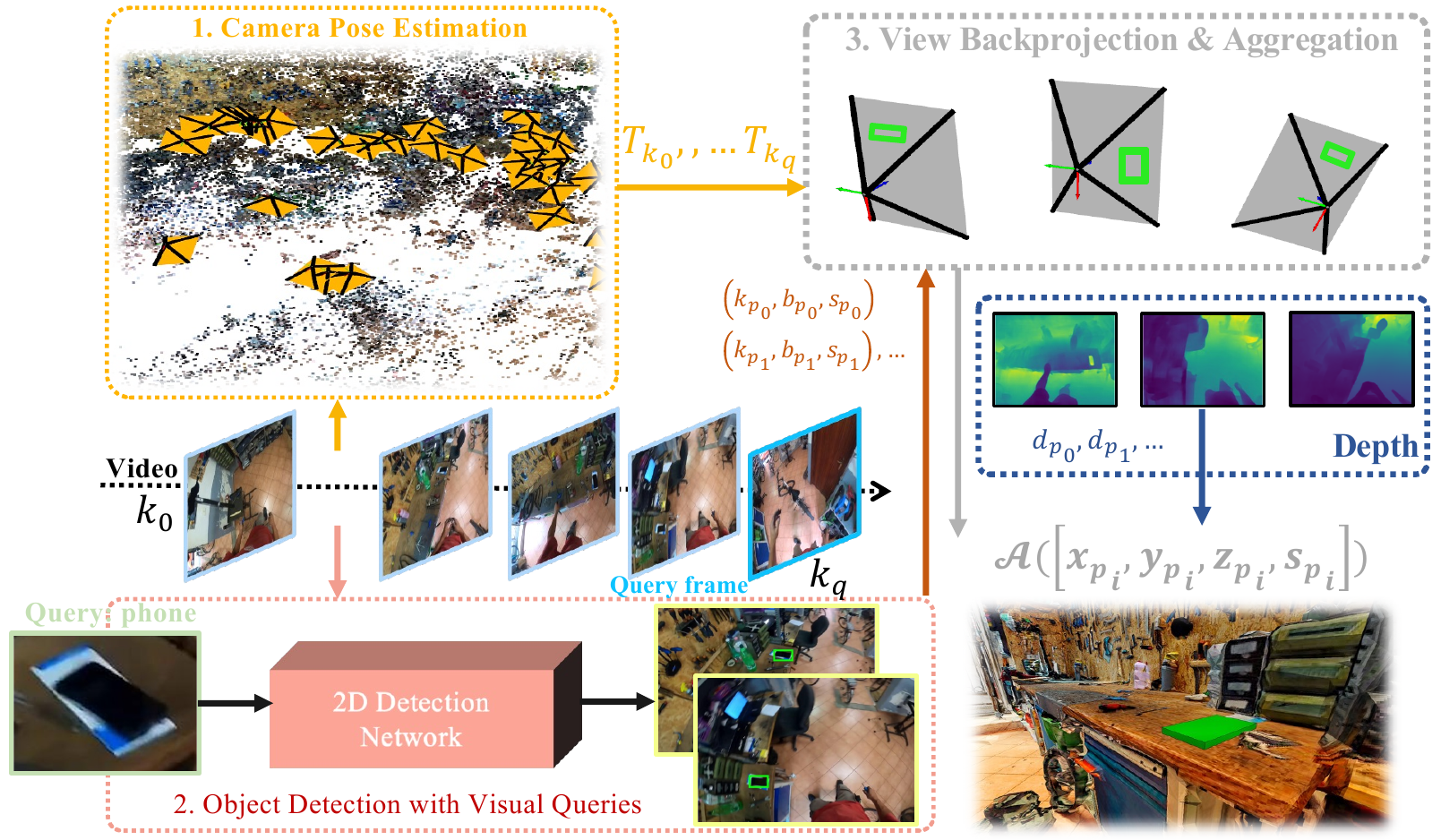}
    \caption{
        \textbf{Main Pipeline.} 
        Our method estimates the camera poses $T$ and retrieves the \textit{query object} from each frame $k_i$ before query frame $k_q$.
        Then, we select the posed frames $k_{p_i}$ with \textbf{peak} 2D response that successfully retrieve the \textit{query object}'s bbox $b_{p_i}$ with high confidence score $s_{p_i}$, estimate the depth $d_{p_i}$ of these frames, and aggregate $\mathcal{A}$ the 3D position $[x_{p_i},y_{p_i},z_{p_i}]$ of the retrieved object to form the final prediction. 
    }
    \label{fig:main}
\end{figure*}

\section{Related Work}

\mysection{2D Detection with Visual Queries} 2D object detection is an essential computer vision task that involves detecting objects within a 2D image and providing their corresponding bounding boxes and class labels.
Over the past few years, deep learning based object detection models~\cite{redmon2016you,girshick2015fast,faster-rcnn,he2015spatial,lin2017focal,tian2019fcos,carion2020end} have achieved remarkable performance on several benchmark datasets~\cite{lin2014microsoft,deng2009imagenet,everingham2009pascal,gupta2019lvis,kuznetsova2020open}. 
However, generalizing trained detectors to unseen classes and the open world~\cite{joseph2021towards,zhao2022revisiting} remains a challenge.
Specially, Visual Queries with 2D Localization (VQ2D)~\cite{grauman2022ego4d}, the sister task of VQ3D,
given a static image crop of the object and an egocentric video recording,
aims to localize the last appearance of the object spatially and temporally,
producing a set of bounding boxes for every frame of a continuous video clip.
VQ2D can be considered as an extension of Few-Shot Detection(FSD)~\cite{antonelli2022few,fan2020few}, where the detection model should be able to quickly transfer to unseen categories given limited new samples.
However, both VQ2D~\cite{xu2022negative,xu2022my} and FSD~\cite{kang2019few,perez2020incremental,yin2022sylph} suffer from false positive detections due to limited positive examples.
In this work, we investigate how to combine VQ2D components specialized for VQ3D.

\mysection{Egocentric Video Understanding} The field of computer vision has witnessed significant advances in understanding third-person view images and videos~\cite{deng2009imagenet,lin2014microsoft,everingham2009pascal,caba2015activitynet,carreira2017quo,karpathy2014large,xu2020g,zhao2021video,alcazar2022end,zhao2023re2tal,vars}. However, the ability to understand visual data from a \textit{first-person} perspective is equally crucial for various research domains, including vision, robotics, and augmented reality. Despite this, it presents unique challenges that require specialized research. In recent years, egocentric vision research has gained substantial attention due to the availability of egocentric datasets~\cite{nguyen2016recognition,sigurdsson2018charades,damen2018scaling,xu2019mo,tome2019xr,akada2022unrealego,grauman2022ego4d}. In response, several studies have been conducted to address the challenges associated with first-person views, such as detecting the camera wearer's hands~\cite{bambach2015lending}, privacy protection~\cite{thapar2021anonymizing}, human-object interactions~\cite{liu20224d}, gaze estimation~\cite{li2018eye}, human body pose estimation~\cite{jiang2017seeing}, and activity recognition and detection~\cite{kazakos2019epic,baradel2018object,li2021ego,kazakos2021little,munro2020multi,ramazanova2023owl}. Particularly noteworthy is Ego4D~\cite{grauman2022ego4d}, which proposes the Visual Query localization task, requiring the agent to locate objects with visual queries in 2D or 3D given recorded egocentric videos.

\mysection{3D Understanding from Egocentric Video} Research on 3D object detection has been extensively conducted using images~\cite{bao2019monofenet,luo2021m3dssd,rukhovich2022imvoxelnet,sparf}, point clouds~\cite{Geiger2012CVPR,song2015sun,pix4point,Li2022sctn} and videos~\cite{huang2018apolloscape,caesar2020nuscenes}. To comprehend and reconstruct 3D scenes from a set of 2D images, Structure from Motion (SfM) \cite{ozyecsil2017survey} has been used. SfM can be categorized into geometric-based methods\cite{schoenberger2016sfm,labbe2019rtab,mur2015orb} that use multiview geometry, learning-based methods~\cite{zhou2017unsupervised,vijayanarasimhan2017sfm,kendall2015posenet} that employ deep neural networks, and hybrid SfM~\cite{teed2018deepv2d,teed2021droid} that combine both approaches. Various SfM methods have been developed to address large-scale videos from dynamic environments~\cite{zhao2022particlesfm} and casual videos from daily life~\cite{zhang2022structure,liu2022depth}.
However, the unique characteristics of egocentric videos, such as dynamics, motion blur, and unusual viewpoints, introduce significant challenges to 3D understanding. Although many studies~\cite{rhodin2016egocap,tome2020selfpose,wang2021estimating, li2023ego, zhang2022egobody,guzov2021human,dai2022hsc4d} have focused on recovering 3D human poses from egocentric videos, very few works have been conducted for egocentric perception in a 3D context. A few notable examples are Chen et al.'s~\cite{chen2022egocentric} study on egocentric indoor localization based on the Manhattan geometry of room layouts, EGO-SLAM~\cite{patra2019ego} that proposed a SLAM system for outdoor egocentric videos using SfM over a temporal window, and NeuralDiff~\cite{tschernezki2021neuraldiff} and N3F~\cite{tschernezki2022neural} that developed a dynamic NeRF from egocentric videos to detect and segment moving objects. Furthermore, Tushar et al.~\cite{nagarajan2022egocentric} proposed an approach that links camera poses and videos to predict human-centric scene context. Our method leverages the 3D structure and egomotion recovered from the egocentric video. It fuses multi-view image detections for enhanced 3D localization of the queried object, which motivates future developments of a strong VQ3D baseline for embodied-AI.

\mysection{Episodic Memory and Embodied AI} Embodied Question Answering (EQA)~\cite{das2018embodied, datta2022episodic}, is a special case of the video-language grounding task, where an embodied agent should answer language questions according to visual observations in 3D indoor environments. 
While EQA usually requires the model to give an answer~\cite{barmann2022did} (\eg language, video clip, \etc) to a language query, VQ3D considers image crops of objects as the query and predicts object displacement as output, which is more intuitive and fundamental for present computer vision techniques. Such a task setting is also strongly related to embodied AI problems~\cite{das2018embodied,chaplot2020learning,anderson2018vision}, but they usually assume the poses are known and operate in stable simulators. Progress in VQ3D has the potential to adapt embodied AI techniques to real-world applications.
\section{Method}

\begin{figure}[t]
    \centering
    \includegraphics[width=0.45\textwidth]{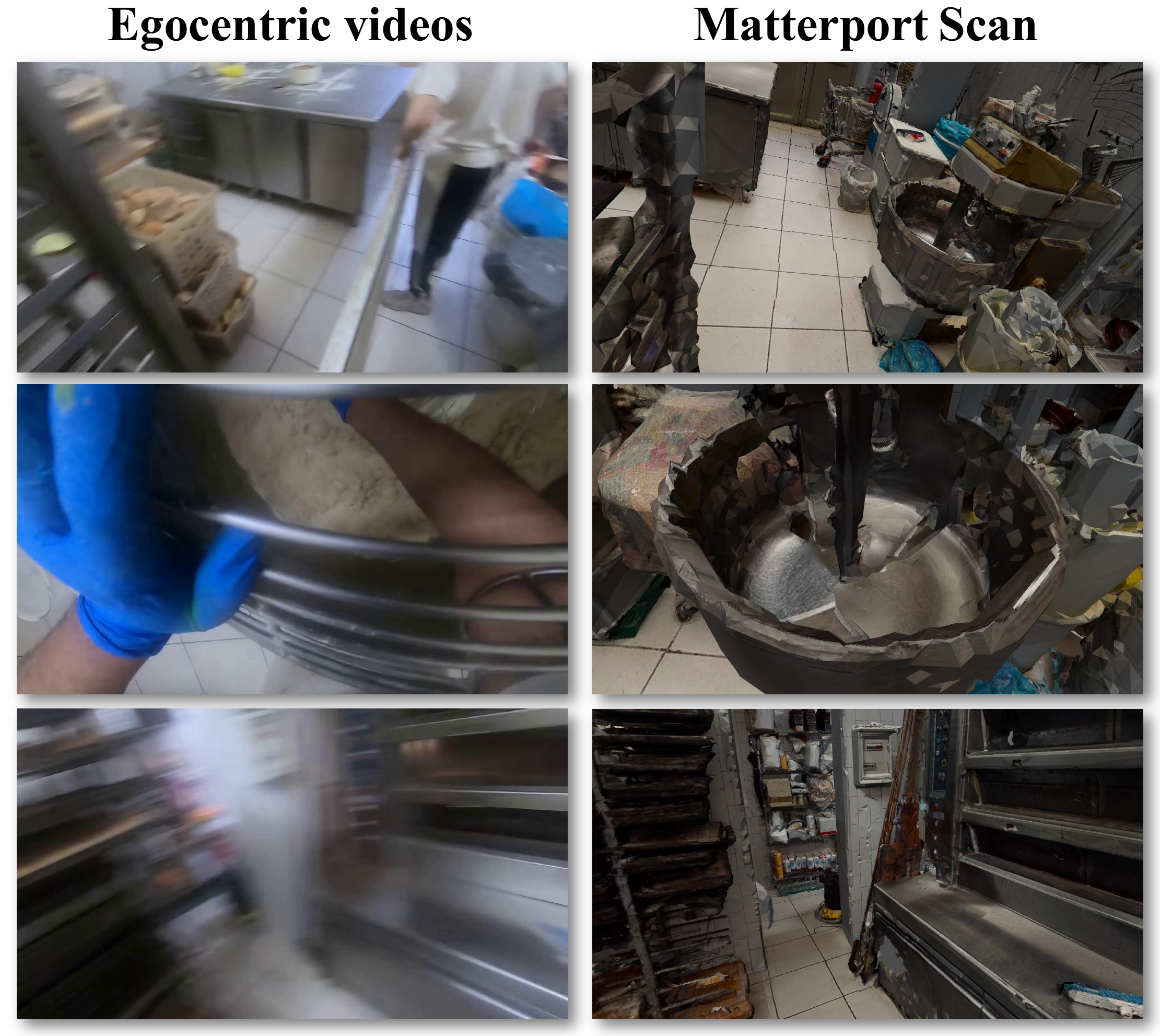}
    \caption{
        \textbf{Egocentric Videos and Matterport Scan.} We show illustrations of the domain gap between Matterport Scan and real scenes. 
        Matterport Scan may have different illumination, scene appearance,
        and missing/low-quality scans.
        Egocentric videos are usually dynamic, free of view (FOV), and have fast motion blur, bringing great challenges for 3D reconstruction and localization. 
    }
    \label{fig:egovideo}
\end{figure}

\subsection{Task and Pipeline Overview}

First, we formalize the VQ3D task as defined in the Ego4D Episodic Memory Benchmark~\cite{grauman2022ego4d}. 
Given an egocentric video $\mathcal{V}$, a query object $o$ defined by a single visual crop $v$, and a query frame $q$, the objective is to estimate the relative displacement vector $\Delta d = (\Delta x, \Delta y, \Delta z)$ defining the 3D location where the query object $o$ was last seen in the environment, with respect to the reference system defined by the 3D pose of the query frame $q$.

To localize a given image query in the video geometrically,
we propose a multi-stage pipeline to entangle 2D information with 3D geometry.
\textbf{First}, we perform Structure from Motion (SfM)~\cite{schoenberger2016sfm}, which estimates the 3D poses $\{T_0, ..., T_{N-1}\}$ for all the $N$ video frames $\{k_{0},...,k_{N-1}\}$. 
\textbf{Second}, we feed the frames of an egocentric video $\mathcal{V}$ and the visual crop $v$ with the query object $o$ to a model that retrieves \textbf{peak} response frames $\{k_{p_0}, k_{p_1}...,  \}$ with corresponding 2D bounding boxes $\{b_{p_0}, b_{p_1}...,  \}$ of the query object $o$. 
\textbf{Finally}, for each response frames $k_{p_i}$, we estimate the depth and back-project the object centroid to 3D using estimated pose $T_{k_{p_i}}$. We recover the world 3D location $[\hat{x},\hat{y},\hat{z}]$ of the object by aggregating per response frame ${p_i}$'s prediction $[x_{p_i},y_{p_i},z_{p_i},s_{p_i}]$.
The final relative displacement vector $\Delta d$ is obtained by projection using $T_q$ with respect to the query frame $q$.
Figure~\ref{fig:main} illustrates an overview of our pipeline.

\subsection{Camera Pose Estimation}

Estimating the camera poses from egocentric videos is difficult, but essential for high-level tasks.
To recover camera poses, the baseline from Ego4D~\cite{grauman2022ego4d} tried to extract the features and match them between the sampled renderings of Matterport Scan and selected frames from the egocentric videos.
However, there is a domain gap between egocentric videos and Matterport scans, as shown in Figure~\ref{fig:egovideo}, and already observed in previous works~\cite{rosano2021embodied,byravan2022nerf2real}, which brings great difficulties for their camera re-localization.
As a result, the Ego4D~\cite{grauman2022ego4d} baseline mostly fails to match video frames, leading to inaccurate 3D reconstruction, low number of Query with Poses (QwP), and low performance in the VQ3D metrics.
To alleviate this issue, we propose to use COLMAP~\cite{schoenberger2016sfm} on the entire sequences and empirically explore the proper hyperparameters for egocentric videos. This leads to an improved QwP rate and improves the overall VQ3D pipeline.

\begin{figure}[t]
    \centering
    \includegraphics[width=\linewidth] {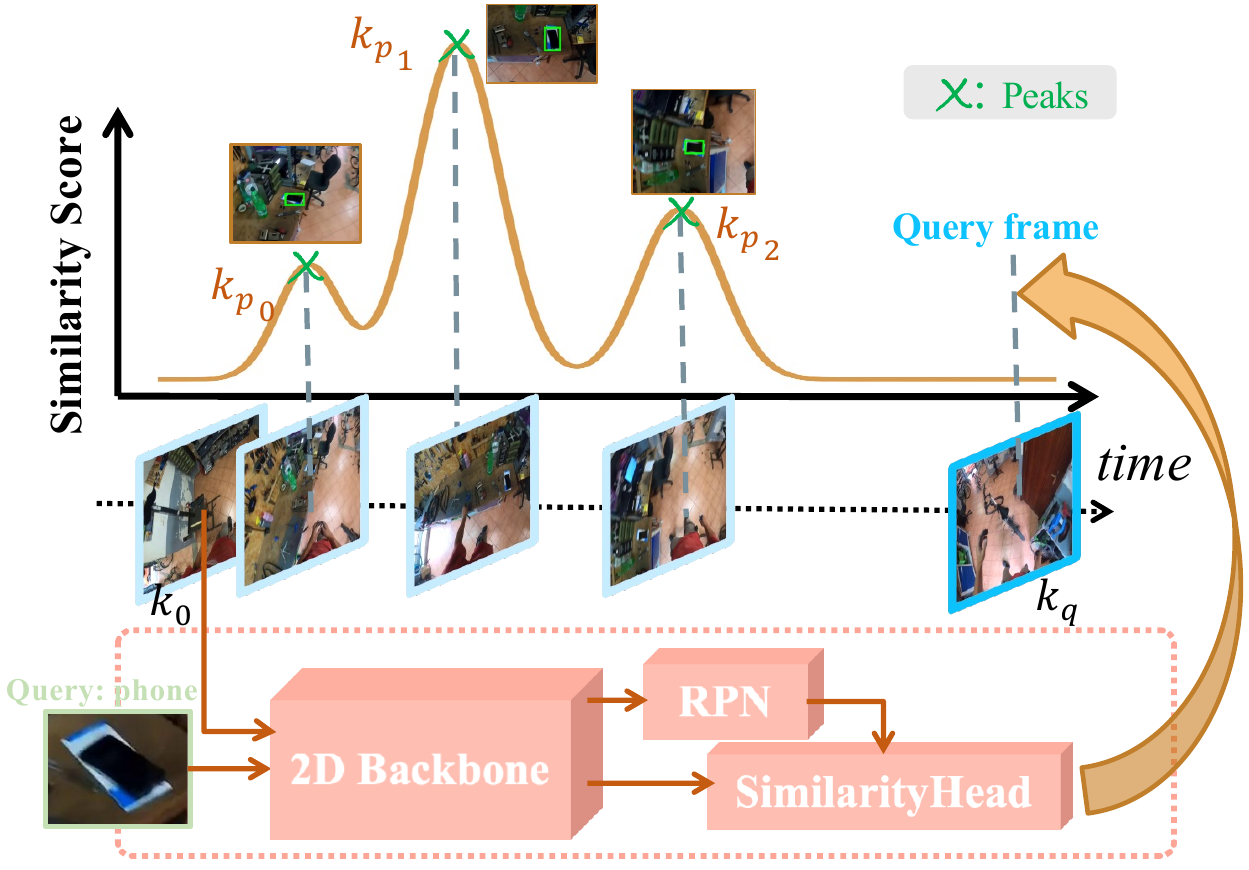}
    \caption{
        \textbf{Our 2D Object Retrieval Module.} The 2D Backbone $\mathcal{F}$ extracts features for both query object $o$ and the input video stream. A pre-trained RPN~\cite{ren2015faster} with ROI-Align~\cite{he2017mask} is then used to generate box proposals and visual features. A Siamese head is trained to evaluate the similarity between query object features and proposal features.
        At inference time, the peak responses in the score signals will be selected as our response frames. 
    }
    \label{fig:2d_det}
\end{figure}

\begin{figure*}[t]
    \centering
    \includegraphics[width=\linewidth] {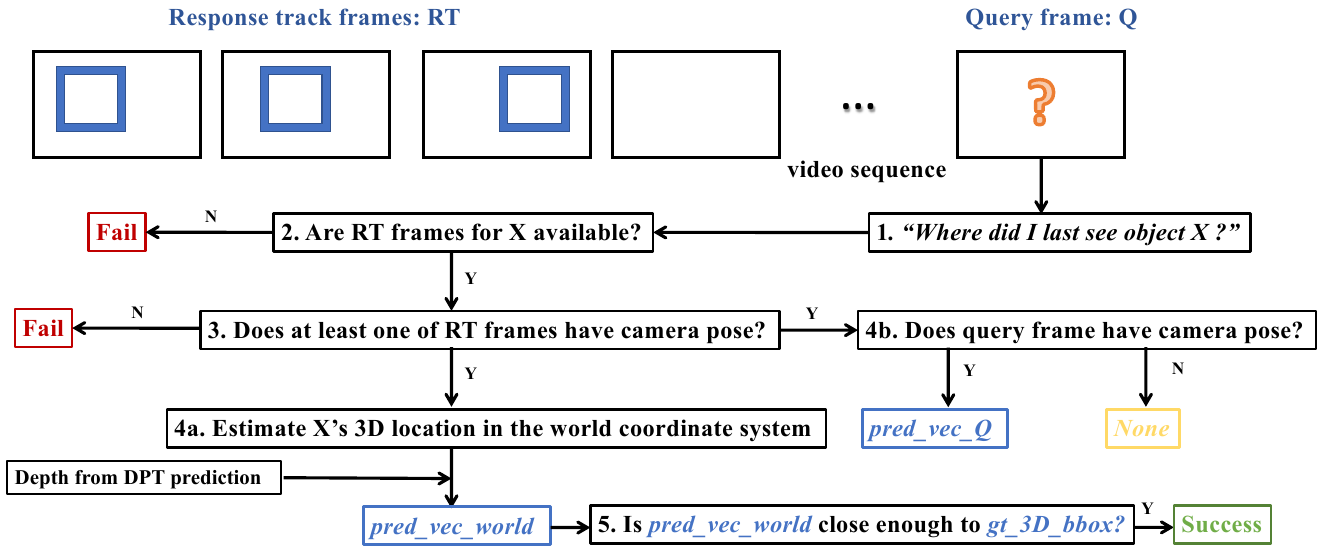}
    \caption{
        \textbf{VQ3D Evluation Metrics.} This flow chart shows how the metrics are calculated for the VQ3D task. The predicted object location in world's coordinate system is $pred\_vec\_world$. The predicted location of the object in the query frame's coordinate system is $pred\_vec\_Q$.
    }
    \label{fig:metric}
\end{figure*}

\subsection{Visual Queries with 2D Localization}
The Video Object Detection from Visual Queries module aims to locate a query object $o$ defined by a single visual crop $v$ both spatially and temporally. The VQ3D baseline proposed by Ego4D~\cite{grauman2022ego4d} builds upon the existing VQ2D pipeline, which includes object detection and tracking stages. However, extending VQ2D to 3D may not be optimal due to the unwieldy and error-prone tracking module, which produces blurry frames and uncertain tracklets that can drift. We propose a revised implementation of VQ2D~\cite{grauman2022ego4d,xu2022negative} that prunes the tracker and integrates the  detector more closely with the 3D modules using a multiview inductive bias to improve the model's reliability and robustness.

\mysection{2D Detection} To detect the query object in a video frame $k_i$, a pre-trained Region Proposal Network (RPN)~\cite{faster-rcnn} with a Feature Pyramid Network (FPN)~\cite{lin2017feature} backbone is utilized to generate a set of bounding box proposals $\{b_{i_0},b_{i_1},...\}$. These proposals are then processed through the RoI-Align operation~\cite{he2017mask} to extract visual features for each box $\{ \mathcal{F}(b_{i_0}),\mathcal{F}(b_{i_1}),...\}$. Meanwhile, we also extract features $ \mathcal{F}(v)$ for the visual crop $v$ using the same FPN backbone. 
To determine whether the query object is present in frame $k_i$, a Siamese head $\mathcal{S}$ is used to output a similarity score between $[0,1]$ for all the bounding box proposals: $\{s_{i_1},s_{i_2}...\}$.
Then we select the Top-1 bounding box proposal as $b_{i}$ for frame $k_i$. 
Finally we can get a tuple of $\{(k_{0},b_{0},s_{0}),...,(k_{q},b_{q},s_{q})\}$ 
for each video frame before the query frame, as shown on the plot at the top of Figure~\ref{fig:2d_det}.

\mysection{Peak Selection}
After smoothing the scores with a median filter, we search for response peaks and select our response frames accordingly.
As the appearance and disappearance of an object in a video can result in a peak signal, we hypothesize that a local peak signal indicates higher confidence for the query object $o$ to be present in the video frame.
Finally, we couple the peak response frames with their corresponding top-1 bounding box proposal and detection similarity score as follows:
$\{(k_{p_0},b_{p_0},s_{p_0}), (k_{p_1},b_{p_1},s_{p_1})...,\}$.

\subsection{Multi-View Unprojection \& Aggregation}

Finally, for each posed response frame, we estimate the 3D position for the center of the 2D bounding box.
We estimate the depth maps for the posed response frames using a pretrained monocular depth estimation network~\cite{ranftl2021vision}.
With the paired response $k_{p_i}$ and the bounding box $b_{p_i}$ together, we retrieve the depth $d_{p_i}$ of the object $o$ as the depth value of its centroid in the depth map.
The backprojection of the 2D position into a 3D displacement vector $[x_{p_i},y_{p_i},z_{p_i}]$ for posed view $k_{p_i}$ is given in Equation~\eqref{eq:backprojection}, where $u_{p_i}, v_{p_i}$ corresponds to the centroid of the bounding box $b_{p_i}$ and $K$ is the intrinsic parameters of the camera estimated by COLMAP.
A visual description of this module is shown on the right side of Figure~\ref{fig:main}.

\begin{align}
     [x_{p_i},y_{p_i},z_{p_i},1]^T  = T_{p_i} d_{p_i} K^{-1}  [u_{p_i},v_{p_i},1]^T
    \label{eq:backprojection}
\end{align}

Unlike the Ego4D baseline~\cite{grauman2022ego4d} that only uses the last tracking response, we propose to aggregate information from multiple points of view to localize the query object $o$,
which leads to a more accurate estimation of the displacement vector $\Delta d$ more robust to noisy posed views.
We make a simple assumption here that recent multiple appearances of an object in a short video clip should be geometrically close to each other in the world coordinate system, in particular, since the object tends to be static without extensive movement~\cite{sener2020temporal,kazakos2021little,wu2019long}.
We defined our aggregation function $\mathcal{A}$ in Equation~\eqref{eq:aggregate}.

\begin{equation}
    [\hat{x},\hat{y},\hat{z}]^T=\mathcal{A}([x_{p_0},y_{p_0},z_{p_0},s_{p_0}]^T,...,[x_{p_i},y_{p_i},z_{p_i},s_{p_i}]^T)
    \label{eq:aggregate}
\end{equation}
To further fuse the temporal predictions across multiview robustly,
we propose taking the 2D detection confidence, \ie the similarity score $s_{p_i}$, into account while fusing multiview. A higher detection score $s_{p_i}$ indicates a higher probability of the presence of a query object at the specific 3D location.
Practically, we use the weighted average as our aggregation operator $\mathcal{A}(\mathbf{x}_{p_i})= \sum s_i \mathbf{x}_{p_i}$.

With the global 3D location of the query object and the camera pose of query frame $T_q$ known, we can finally give the prediction for relative displacement vector:

\begin{equation}
    \Delta \hat{d}=T_q^{-1}[\hat{x},\hat{y},\hat{z},1]^T
    \label{eq:displacement}
\end{equation}

\section{Experiments}
\subsection{Ego4D VQ3D Benchmark}
Ego4D~\cite{grauman2022ego4d} is a massive-scale egocentric video dataset and benchmark suite.
As a part of the episodic memory benchmark, the VQ3D task contains 164, 44, and 69 video clips for train, validation, and test set, respectively.
There are 164 and 264 visual queries in the validation and test sets, respectively. 
The video clips typically last from 5 to 10 minutes.
All those video clips were recorded in indoor scenes that have been scanned with Matterport devices before.
The ground-truth 3D locations of the objects were annotated by volunteers in the Matterport scans.

\subsection{Experimental Setup}
To be robust to motion blur, we subsample 100 contiguous non-blurry frames from the videos selected by the variance of Laplacian greater than 100, which will be fed into COLMAP auto-reconstruction to estimate camera intrinsic at first.
To model the fisheye distortion in egocentric videos, we choose the \textit{RADIAL\_FISHEYE} camera.
We then set the sequential matcher with $window\_size=10$ in COLMAP sparse reconstruction for the whole video.
Since the poses from SfM have the \textit{scale ambiguity}~\cite{hartley2003multiple} issue, we align the COLMAP reconstruction to Matterport scan coordinate system as post-processing. 
We render at least three images from the Matterport scan with known camera poses and then perform a $Sim3$ transformation to align the COLMAP coordinate system.
The 2D Faster-RCNN~\cite{faster-rcnn} backbone is pretrained on MS-COCO~\cite{coco} and frozen. We only train the Siamese head in VQ2D~\cite{grauman2022ego4d} train set. 
We perform a median filter with $kernel\_size=5$ frames and select the detection peaks~\cite{scipy_peak} supported by $peak\_width\geq 3, distance\geq 25$ from the detection score curve.
We adopt the DPT~\cite{ranftl2021vision} network pretrained on NYU V2~\cite{Silberman:ECCV12} for depth estimation.

\subsection{Metrics}
For a fair comparison, we use the same metrics as defined by Ego4D's baseline~\cite{grauman2022ego4d}.
Given the predicted 3D location and the ground truth object 3D location,
\textbf{Angle} corresponds to its angular error in radians, and \textbf{L2} corresponds to its distance error in meters using the root mean square error (RMSE).
\textbf{QwP} represents the query ratio for which we have pose estimation for both the response frames and the query frame. 
The \textbf{Success} metric is formulated as the ratio of query prediction, whose \textbf{L2} error is smaller than a threshold, to total queries. Note that if we fail to estimate the camera pose for the response frames or the query frame,  the corresponding query will be considered as failed directly.
The \textbf{Success$^*$} means the success metric computed only for queries with associated pose estimates.

In summary, we provide a detailed flow chart to explain how the evaluation works in Figure~\ref{fig:metric}. The metrics can be hence calculated as follows.

$$
    QwP=\frac{ \text{Queris with both } pred\_vec\_world, pred\_vec\_Q }{\text{Total queries in Step 1}}
    $$
 $$
     Succ^*\ Rate=\frac{ \text{Successful queries in Step 5} }{ \text{Total queries with RT camera poses in Step 4a}}, 
     $$
 $$
     Succ\ Rate=\frac{ \text{Successful queries in Step 5} }{ \text{Total queries in Step 1}}
     $$

\subsection{Our Results}

\begin{table}[t]
\setlength{\tabcolsep}{3pt}
\centering
\resizebox{0.99\linewidth}{!}{%
\begin{tabular}{@{}l|ccccc@{}}
\toprule
\multicolumn{1}{c}{\textbf{Method}} & \multicolumn{5}{c}{\textbf{Validation Set}} \\
 & \textbf{Succ\%↑} & \textbf{Succ*\%↑} & \textbf{L2↓} & \textbf{Angle↓} & \textbf{QwP\%↑} \\ \midrule
Ego4D~\cite{grauman2022ego4d}                & 1.22    & 30.77   & 5.98   & 1.6   & 1.83   \\
Ego4D$^*$             & 73.78   & 91.45   & 2.05   & 0.82  & 80.49  \\
\textbf{EgoLoc (ours)}                 & \textbf{80.49}   & \textbf{98.14}   & \textbf{1.45}   & \textbf{0.61}  & \textbf{82.32}  \\ \midrule
\multicolumn{1}{l}{} & \multicolumn{5}{c}{\textbf{Test Server Leaderboard}}       \\ \midrule
Ego4D~\cite{grauman2022ego4d}                & 8.71    & 51.47   & 4.93   & 1.23  & 15.15  \\
Ego4D$^*$             & 77.27   & 86.06   & 2.37   & 1.14  & 90.15  \\
\textbf{EgoLoc (ours)}                 & \textbf{87.12}   & \textbf{96.14}   & \textbf{1.86}   & \textbf{0.92}  & \textbf{90.53} \\ \bottomrule
\end{tabular}%
}
\vspace{2pt}
\caption{
        \textbf{Main Results.}
        We show our results on the validation and test sets of the VQ3D task from the Ego4D Episodic Memory Benchmark. We compare against Ego4D \cite{grauman2022ego4d} baseline and 
        $Ego4D^*$, an improved baseline by just replacing the camera pose estimation part to ours. %
    }
    \label{tb:main}
\end{table}

We present the main results of our study in Table~\ref{tb:main}, where we demonstrate significant improvements in both the validation and test sets across all metrics. As we had hypothesized, the low success rate in the Ego4D dataset was largely due to the low Queries with Poses (QwP) ratio, which serves as the upper bound for the overall success rate. Without frame poses, it is impossible to estimate 3D displacement. The low QwP ratio in Ego4D was caused by their camera pose estimation method, which tried to perform feature matching across a large domain gap between simulated and real scenes, as can be seen in the first row of our results in Table~\ref{tb:main}. Our baseline of the new proposed pipeline Ego4D$^*$, by just replacing the camera pose estimation part of Ego4D, achieved a success rate of $73.78\%$ and $77.27\%$ in the validation and test sets, respectively, compared to Ego4D's $1.22\%$ and $8.71\%$.

Despite similar QwP values, our proposed aggregation method demonstrates significant enhancements to the improved Ego4D$^*$ baseline in terms of overall success, L2 accuracy, and angle accuracy. The use of multi-view fusion significantly improves localization accuracy, as evidenced by the reduction in L2 error. These improvements translate to a 6.71\% increase in overall success rate on the validation set and a 9.85\% increase on the test set.

\begin{figure*}[t]
    \centering
    \includegraphics[width=0.95\linewidth] {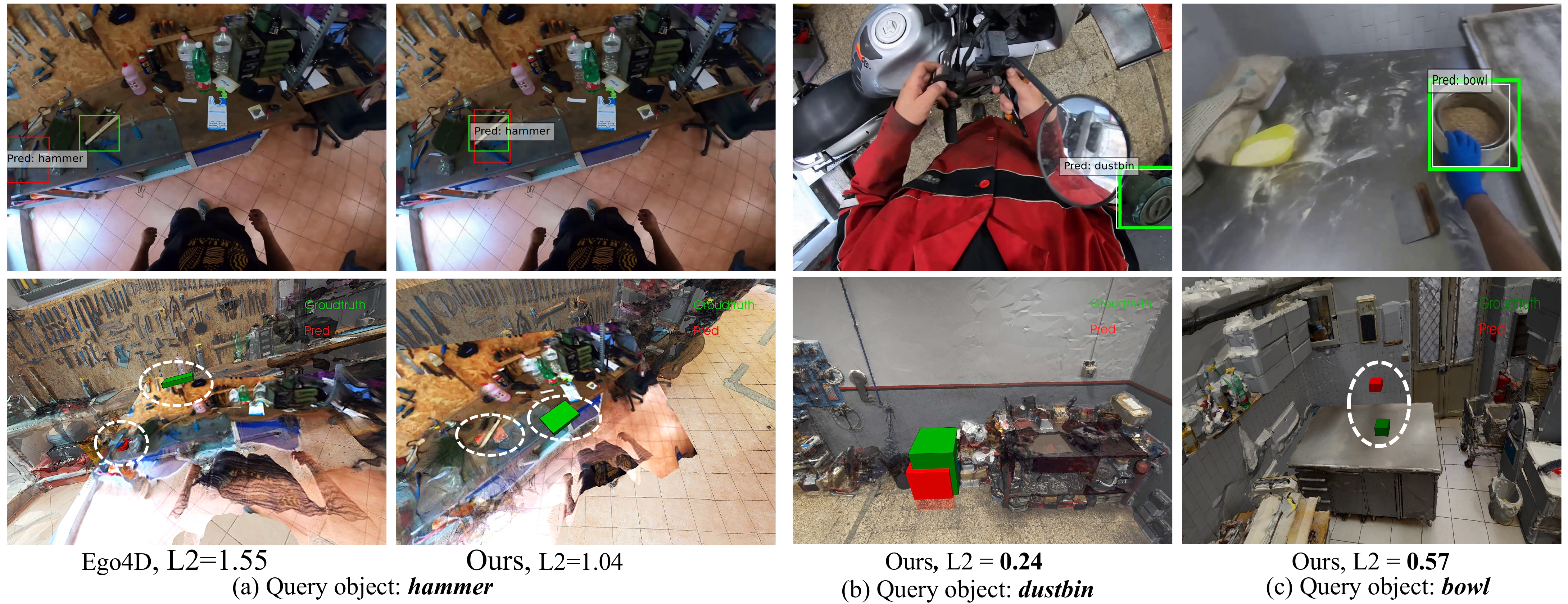}
    \caption{ \textbf{Visualization of 2D responses and 3D Localization.}
        We visualize the 2D response detection on the top row with the corresponding 3D localization below. We backproject the response frame to an example Matterport scan using estimated depth map and camera poses to show a superposition. The highlighted area is backprojection, and the faded area is groundtruth scene scan. \textbf{For query (a)}, our object detection selection gives the correct bbox while the tracking result from Ego4D has shifted, leading to better 3D localization. 
        \textbf{For query (b) }, our method shows pretty good localization accuracy, even though the view angle is quite unusual along with severe fisheye distortion. 
        \textbf{For query (c)}, since the wearer is grabbing and moving the bowl, our prediction is shifted a little bit.
    }
    \label{fig:vis_final}
\end{figure*}

\section{Analysis}
We conduct an extensive empirical analysis of different modules on the validation set. 
\subsection{Ablation Study}

\begin{table}[t]
\setlength{\tabcolsep}{3pt}
\centering
\resizebox{0.96\linewidth}{!}{%
\begin{tabular}{@{}l|ccccc@{}}
\toprule
\textbf{VQ2D} & \textbf{Succ\%↑} & \textbf{Succ*\%↑} & \textbf{L2↓} & \textbf{Angle↓} & \textbf{QwP\%↑} \\ \midrule
LastTrack   & 73.78 & 91.45 & 2.05 & 0.82 & 80.49 \\
LastDetPeak & 78.05 & 94.41 & 1.81 & 0.7  & 82.32 \\
TopDetPeak  & 79.88 & 96.89 & 1.48 & \textbf{0.51} & 82.32 \\
\textbf{DetPeaks }    & \textbf{80.49} & \textbf{98.14} & \textbf{1.45} & 0.61 & 82.32 \\ \bottomrule
\end{tabular}%
}
\vspace{2pt}
\caption{
        \textbf{Ablation study on Detection}
        We ablate on how to use VQ2D results, comparing selecting the last frame of tracking,
        last detection peak or the highest detection response peak. Our EgoLoc pipeline utilizes multiple detection peaks for aggregation.
    }
    \label{tb:det}
\end{table}
\mysection{VQ2D}  
Table~\ref{tb:det} presents the findings of the analysis conducted on 2D object detection from egocentric videos on the VQ3D performance. To localize the object in 2D, Ego4D employs the ``Detection and Tracking'' strategy, specifically ``LastTrack''. This strategy involves selecting the detection response peak closest to the query frame (Figure~\ref{fig:2d_det}), running the tracker forward and backward to obtain the response track, and using the last frame from the response track as the answer to the question ``where did I last see object X?''. However, this approach has limitations, including time and computation overheads, blurry response, objects being far from the center of the frame, and difficulties in depth estimation for backprojection to 3D.

Our study focuses more on the confidence and precision of the question ``where did I \textbf{last} see object X?''. To this end, the ``LastTrack'' approach is compared to ``LastDetPeak'' (selecting the last peak), ``TopDetPeak'' (selecting the highest peak), and ``DetPeaks'' (selecting multi-peak responses motivated by multiview aggregation). The results indicate that even considering one peak can enhance the ``LastTrack'' approach from $73.78\%$ to $78.05\%$. Furthermore, our ``DetPeaks'' outperforms the other approaches, achieving a Top-1 performance of over $80\%$ in overall success.

\begin{table}[t]
\setlength{\tabcolsep}{5pt}
\centering
\resizebox{0.97\linewidth}{!}{%
\begin{tabular}{l|ccc} 
\toprule
\textbf{VQ2D} & \textbf{Succ*\%↑} & \textbf{L2↓} & \textbf{Angle↓} \\ \midrule
Last GT Track      & 98.04             & 1.22         & \textbf{0.41}            \\
Mean GT Track      & \textbf{98.69}             & \textbf{1.17}         & \textbf{0.41}            \\
EgoLoc (on GT 2D Track)     & 97.39             & 1.47         & 0.64           \\ \bottomrule
\end{tabular}%
}
\vspace{2pt}
\caption{
        \textbf{Compared to GT 2D Tracking.}
        We compare our results based on the ground truth 2D annotation, i.e., ground truth tracking result for the last appearance of the query object. *Since different 2D response frames will affect the QwP ratio, we only compare the queries where both GT response frames and detection peak frames have camera poses to ensure only 2D detection accuracy will influence the relative localization accuracy. 
    }
    \label{tb:gt2d}
\end{table}

\mysection{VQ2D GT}
Furthermore, the VQ3D benchmark, which is a subset of VQ2D data, includes ground truth tracking annotations for the query object's last appearance. To evaluate the pipeline, peak responses are substituted as ground truth tracking since ground truth (GT) does not have scores from detection. The study ablates on the last GT frame and multi-view mean of GT frame strategy. The results, as presented in Table~\ref{tb:gt2d}, demonstrate that ground truth 2D annotations can decrease the object localization shift, with reduced L2 and angular error. However, even though it achieves the highest success rate, the overall success rate is only $0.61\%$ higher than that of Detection Peaks. This finding indicates that the 2D object retrieval module has almost reached saturation compared to the camera pose estimation. Although the current Siamese detector still lacks 2D detection accuracy~\cite{grauman2022ego4d,xu2022negative}, we are approaching the upper bound of potential improvements gained from 2D object retrieval. Because 2D metrics in pixels, e.g., \textit{IoU}, tend to be stricter than current 3D metrics in meters.

\begin{table}[!htb]
\setlength{\tabcolsep}{4pt}
\centering
\resizebox{0.99\linewidth}{!}{%
\begin{tabular}{l|ccccc}
\toprule
 \multicolumn{1}{c}{\textbf{View}} &   &   &   &   & \\
 \multicolumn{1}{c}{\textbf{Aggregation}}  & \textbf{Succ\%↑} & \textbf{Succ*\%↑} & \textbf{L2↓} & \textbf{Angle↓} & \textbf{QwP\%↑} \\\midrule
Last        & 78.05 & 94.41 & 1.81 & 0.7  & 82.32 \\
Mean        & 79.88 & 97.52 & 1.53 & 0.66 & 82.32 \\
NMS         & 79.88 & 96.89 & \textbf{1.43} & \textbf{0.5}  & 82.32 \\
\textbf{Det Weighted} &\textbf{80.49} & \textbf{98.14} & 1.45 & 0.61 & 82.32 \\
\bottomrule
\end{tabular}%
}
\vspace{2pt}
\caption{
        \textbf{Ablation Study on Multi-View Aggregation}
        We ablate different ways to aggregate the 3D prediction from multi-view 2D responses. Weighted aggregation based on detection confidence is what EgoLoc pipeline utilizes 
    }
    \label{tb:agg}
\end{table}

\mysection{Multiview Aggregation} 
The simplest way to do backprojection for object $o$ in 3D is taking one single response for Equation~(\ref{eq:backprojection}).
However, since our pipeline is multi-stage, prediction from one single frame could be fragile to the errors we accumulated in each module. Aggregating 3D displacements predicted from multiview strengthens the 3D precision.

In Table~\ref{tb:agg}, given the response peaks, we ablate on different multiview aggregation functions $\mathcal{A}$. ``Mean'' is the most straightforward and naive way where we set $\mathcal{A}=mean$. ``DetWeighted'' is our proposed method to fuse 3D predictions from multiview peaks based on 2D object retrieval confidence. ``Non Maximum Suppression (NMS)'' is inspired by the 2D detection field~\cite{girshick2014rich} where we select the 3D prediction point with the highest confidence score and fuse it with the neighborhood points also by the confidence score. In this way, the predicted locations with low confidence and far away from Top-1 won't be considered, but the final prediction still benefits from multiview.

 As we have found from Table~\ref{tb:agg}, Mean and NMS give a comparable performance. But the best performance, a boost of $2.44\%$, is observed from the proposed detection confidence fusion.

\begin{table}[!thb]
\setlength{\tabcolsep}{4pt}
\centering
\resizebox{0.97\linewidth}{!}{%
\begin{tabular}{l|ccccc}
\toprule
\textbf{Triangulation} & \textbf{Succ\%↑} & \textbf{Succ*\%↑} & \textbf{L2↓} & \textbf{Angle↓} & \textbf{QwP\%↑} \\ \midrule
DetPeaks & 56.1  & 67.7  & 6.2  & 1.16 & \textbf{82.32} \\
TrackGT       & 56.71 & 71.9  & 4.88 & 1.26 & 81.1  \\
DPT              & \textbf{80.49} & \textbf{98.14} & \textbf{1.45} & \textbf{0.61} & \textbf{82.32} \\ \bottomrule
\end{tabular}%
}
\vspace{2pt}
\caption{
        \textbf{Ablation study on Depth Estimation}
        We try to do N-view triangulation given the 2D responses for the same 3D object across multiple known posed images. TrackGT is the 2D tracking ground truth annotation for the last appearance. DPT~\cite{ranftl2021vision} is used in EgoLoc's triangulation.
    }
    \label{tb:depth}
\end{table}

\mysection{Depth Estimation} We adopt the off-the-shelf monocular depth estimation network, DPT~\cite{ranftl2021vision}, robust to blurry and fisheye distortion, to get a smooth depth estimation for the object. However, since we have paired 2D points with camera poses $\{(k_{p_0},b_{p_0},T_{p_0}), (k_{p_1},b_{p_1},T_{p_1})...,\}$ for peak frames,
it's also possible to estimate the depth $d$ from multi-view geometry by triangulation~\cite{hartley2003multiple} if we have detected more than one peak response. This can be done by solving the combined equation system consisting of Equation~\ref{eq:backprojection} from each view.

At first, we use the camera intrinsic estimated by COLMAP to undistort the images.
Then, we have tried to use our detected response peaks and groundtruth 2D tracking to triangulate. 
As shown in Table~\ref{tb:depth}, we find a catastrophic drop in L2, angle, and overall performance.
Even triangulation from groundtruth 2D bboxes is just slightly better than using detection peaks but still has a massive localization L2 error as large as $4.88$. 

We think the reason comes two-fold: 1) Triangulation, as a naive approach, is sensitive to the quality of undistortion and view consistency, while the distortion coefficients and camera poses estimated by COLMAP are not precise enough; 2) Also, the short baseline due to the small translation with fast rotation commonly seen in egocentric videos brings extra difficulties for a numerically stable solution.

\mysection{Complete Ablations} We present a more complete and unified ablation in Table~\ref{tb:test}. It shows that each module provides an improvement over Ego4D baseline on test set. Though the most significant gain is brought by our adaption of Egocentric SfM, our other modules also have nontrivial contributions to further improve by about 10\%.
Our final model \textit{MV-Peaks Weighted} uses the weighted average predictions from confident peak detections along with egocentric SfM.

\begin{table}[t]
\setlength{\tabcolsep}{2pt}
\centering
\resizebox{0.99\linewidth}{!}{%
\begin{tabular}{@{}l|ccccc@{}}
\toprule
\multicolumn{1}{c}{\textbf{Method}} & \multicolumn{5}{c}{\textbf{Test Server Leaderboard}} \\
 & \textbf{Succ\%↑} & \textbf{Succ*\%↑} & \textbf{L2↓} & \textbf{Angle↓} & \textbf{QwP\%↑} \\ \midrule
Ego4D baseline               & 8.71    & 51.47   & 4.93   & 1.23  & 15.15  \\
+Egocentric SfM             & 77.27   & 86.06   & 2.37   & 1.14  & 90.15  \\
+Last Detection & 81.06   & 90.35   & 2.20  & 1.24  & 90.53  \\
+Detection Peak & 85.61   & 94.98   & 1.85  & 1.24  & 90.53  \\
+MV-Peaks Average & 86.36   & 95.37   & 1.94  & 1.19  & 90.53  \\
\textbf{+MV-Peaks Weighted }                 & \textbf{87.12}   & \textbf{96.14}   & \textbf{1.86}   & \textbf{0.92}  & \textbf{90.53} \\ \bottomrule
\end{tabular}%
}
\vspace{2pt}
\caption{
        \textbf{Full Ablations.}
        We list the performance of adding each module 
        in EgoLoc compared to the Ego4D \cite{grauman2022ego4d} baseline. 
        MV refers to multi-view. EgoLoc utilizes weighted multi-view aggregation  based on multiple detection peaks (\textbf{+MV-Peaks Weighted }).
    }
    \label{tb:test}
\end{table}

\mysection{Scene Variance} \label{sec:scene} We notice that the absolute success rate in validation and test set has big differences, comparing Ego4D's 1.22\% with 8.71\%, our 77.44\% with 86.36\% in Table~\ref{tb:main}. 
So we ablate our method's performance across different scenes in Table~\ref{tb:scene}.
We find that the QwP is dominating the overall performance, especially for the hard scene, \textit{Bakery}, where COLMAP only gets poses for $44.12\%$ of queries. Also, the variance of L2 in different scenarios is also significant.
This implies that egocentric videos recorded in different scenes and activities may have very different characteristics, making the VQ3D task more challenging.

\mysection{Visualization} We provide the visualization results for some queries in Figure~\ref{fig:vis_final}. For query (a), Ego4D's detection and tracking strategy does not perform well. It may detect the wrong object at one previous frame and keep tracking it, or the tracker may lose track of the correct object. Our method, instead, considers multiview peaks with confidence, thus providing more accurate object recognition and localization. However, query (c) also shows a flaw in our assumption that the object tends to be static across multiple peak responses. Here since the wearer is interacting with the bowl, the fusion from other views shifts the weighted average prediction floating above the table.
\begin{table}[t]
\centering
\resizebox{0.99\linewidth}{!}{%
\tabcolsep=0.13cm
\begin{tabular}{l|ccccc}
\toprule
\textbf{Scene} & \textbf{Succ\%↑} & \textbf{Succ*\%↑} & \textbf{L2↓} & \textbf{Angle↓} & \textbf{QwP\%↑} \\ \midrule
All                        & 80.49 & 98.14 & 1.45 & 0.61 & 82.32 \\
Scooter mechanic 31 & 91.89 & 97.3  & 1.35 & 0.5  & 94.59 \\
Baker 32            & 41.18 & 96.77 & 1.68 & 0.76 & 44.12 \\
Carpenter 33        & 82.14 & 100   & 1.48 & 1    & 82.14 \\
Bike mechanic 34    & 96.43 & 100   & 1.43 & 0.48 & 96.43 \\ \bottomrule
\end{tabular}%
}
\vspace{2pt}
\caption{
        \textbf{Variance of Egocentric Scenes.}
       We show EgoLoc's results on some scenes from Ego4D \cite{grauman2022ego4d}. Different scenes have completely different room layouts, lighting, and human activities,. This is part of the challenge of VQ3D.
    }
    \label{tb:scene}
\end{table}

\subsection{Discussions and Insights}

Our investigation into VQ3D has revealed several key issues, highlighting avenues for further research. Our camera pose estimation remains a bottleneck, and developing robust SfM/SLAM algorithms for dynamic egocentric videos is a crucial next step. Furthermore, an end-to-end learning-based solution for camera and object re-localization could have applications in online settings, such as wearable AI assistants. Constructing a 4D episodic memory of the dynamic 3D environment is another promising direction. Our work represents a new starting point for further research in egocentric 3D understanding. Ultimately, we hope that our work will serve as a new starting point for further research in this field. And we believe that further investigation in VQ3D could yield interesting and valuable insights toward egocentric 3D understanding.

\section{Conclusions and Future Works}

In this work, we have presented a reformulation of the VQ3D task and a modular pipeline that leads to significant improvements on the Ego4D VQ3D benchmark. Through numerous experiments and ablations, we have validated our proposed methodology and demonstrated its effectiveness. Nevertheless, we recognize that this is just the first step towards addressing the challenges of 4D understanding and we anticipate further research in this direction.
Additionally, our successful methods and strategies may be applied to classical vision tasks in video understanding, offering new possibilities for leveraging 3D knowledge in such settings. We believe that these developments will contribute to a better understanding of complex dynamic scenes and help unlock new applications in fields such as robotics, autonomous driving, and augmented/virtual reality.

\mysection{Acknowledgement}\label{sec:ack}
The authors would like to thank Guohao Li, Jesus Zarzar and Sara Rojas Martinez for the insightful discussion. This work was supported by the KAUST Office of Sponsored Research through the Visual Computing Center funding, as well as, the SDAIA-KAUST Center of Excellence in Data Science and Artificial Intelligence (SDAIA-KAUST AI). Part of the support is also coming from the KAUST Ibn Rushd Postdoc Fellowship program.   
{\small
\typeout{}
\bibliographystyle{ieee_fullname}
\bibliography{egbib}

\begin{thebibliography}{100}\itemsep=-1pt

\bibitem{akada2022unrealego}
Hiroyasu Akada, Jian Wang, Soshi Shimada, Masaki Takahashi, Christian Theobalt,
  and Vladislav Golyanik.
\newblock Unrealego: A new dataset for robust egocentric 3d human motion
  capture.
\newblock In {\em Computer Vision--ECCV 2022: 17th European Conference, Tel
  Aviv, Israel, October 23--27, 2022, Proceedings, Part VI}, pages 1--17.
  Springer, 2022.

\bibitem{alcazar2022end}
Juan~Leon Alcazar, Moritz Cordes, Chen Zhao, and Bernard Ghanem.
\newblock End-to-end active speaker detection.
\newblock In {\em Proceedings of the European Conference on Computer Vision
  (ECCV)}, 2022.

\bibitem{anderson2018vision}
Peter Anderson, Qi Wu, Damien Teney, Jake Bruce, Mark Johnson, Niko
  S{\"u}nderhauf, Ian Reid, Stephen Gould, and Anton Van Den~Hengel.
\newblock Vision-and-language navigation: Interpreting visually-grounded
  navigation instructions in real environments.
\newblock In {\em Proceedings of the IEEE conference on computer vision and
  pattern recognition}, pages 3674--3683, 2018.

\bibitem{antonelli2022few}
Simone Antonelli, Danilo Avola, Luigi Cinque, Donato Crisostomi, Gian~Luca
  Foresti, Fabio Galasso, Marco~Raoul Marini, Alessio Mecca, and Daniele
  Pannone.
\newblock Few-shot object detection: A survey.
\newblock {\em ACM Computing Surveys (CSUR)}, 54(11s):1--37, 2022.

\bibitem{bambach2015lending}
Sven Bambach, Stefan Lee, David~J Crandall, and Chen Yu.
\newblock Lending a hand: Detecting hands and recognizing activities in complex
  egocentric interactions.
\newblock In {\em Proceedings of the IEEE international conference on computer
  vision}, pages 1949--1957, 2015.

\bibitem{bao2019monofenet}
Wentao Bao, Bin Xu, and Zhenzhong Chen.
\newblock Monofenet: Monocular 3d object detection with feature enhancement
  networks.
\newblock {\em IEEE Transactions on Image Processing}, 29:2753--2765, 2019.

\bibitem{baradel2018object}
Fabien Baradel, Natalia Neverova, Christian Wolf, Julien Mille, and Greg Mori.
\newblock Object level visual reasoning in videos.
\newblock In {\em Proceedings of the European Conference on Computer Vision
  (ECCV)}, pages 105--121, 2018.

\bibitem{barmann2022did}
Leonard B{\"a}rmann and Alex Waibel.
\newblock Where did i leave my keys?-episodic-memory-based question answering
  on egocentric videos.
\newblock In {\em Proceedings of the IEEE/CVF Conference on Computer Vision and
  Pattern Recognition}, pages 1560--1568, 2022.

\bibitem{byravan2022nerf2real}
Arunkumar Byravan, Jan Humplik, Leonard Hasenclever, Arthur Brussee, Francesco
  Nori, Tuomas Haarnoja, Ben Moran, Steven Bohez, Fereshteh Sadeghi, Bojan
  Vujatovic, et~al.
\newblock Nerf2real: Sim2real transfer of vision-guided bipedal motion skills
  using neural radiance fields.
\newblock {\em arXiv preprint arXiv:2210.04932}, 2022.

\bibitem{caba2015activitynet}
Fabian Caba~Heilbron, Victor Escorcia, Bernard Ghanem, and Juan Carlos~Niebles.
\newblock Activitynet: A large-scale video benchmark for human activity
  understanding.
\newblock In {\em Proceedings of the ieee conference on computer vision and
  pattern recognition}, pages 961--970, 2015.

\bibitem{caesar2020nuscenes}
Holger Caesar, Varun Bankiti, Alex~H Lang, Sourabh Vora, Venice~Erin Liong,
  Qiang Xu, Anush Krishnan, Yu Pan, Giancarlo Baldan, and Oscar Beijbom.
\newblock nuscenes: A multimodal dataset for autonomous driving.
\newblock In {\em Proceedings of the IEEE/CVF conference on computer vision and
  pattern recognition}, pages 11621--11631, 2020.

\bibitem{carion2020end}
Nicolas Carion, Francisco Massa, Gabriel Synnaeve, Nicolas Usunier, Alexander
  Kirillov, and Sergey Zagoruyko.
\newblock End-to-end object detection with transformers.
\newblock In {\em Computer Vision--ECCV 2020: 16th European Conference,
  Glasgow, UK, August 23--28, 2020, Proceedings, Part I 16}, pages 213--229.
  Springer, 2020.

\bibitem{carreira2017quo}
Joao Carreira and Andrew Zisserman.
\newblock Quo vadis, action recognition? a new model and the kinetics dataset.
\newblock In {\em proceedings of the IEEE Conference on Computer Vision and
  Pattern Recognition}, pages 6299--6308, 2017.

\bibitem{chaplot2020learning}
Devendra~Singh Chaplot, Dhiraj Gandhi, Saurabh Gupta, Abhinav Gupta, and Ruslan
  Salakhutdinov.
\newblock Learning to explore using active neural slam.
\newblock {\em arXiv preprint arXiv:2004.05155}, 2020.

\bibitem{chen2022egocentric}
Xiaowei Chen and Guoliang Fan.
\newblock Egocentric indoor localization from coplanar two-line room layouts.
\newblock In {\em Proceedings of the IEEE/CVF Conference on Computer Vision and
  Pattern Recognition}, pages 1549--1559, 2022.

\bibitem{dai2022hsc4d}
Yudi Dai, Yitai Lin, Chenglu Wen, Siqi Shen, Lan Xu, Jingyi Yu, Yuexin Ma, and
  Cheng Wang.
\newblock Hsc4d: Human-centered 4d scene capture in large-scale indoor-outdoor
  space using wearable imus and lidar.
\newblock In {\em Proceedings of the IEEE/CVF Conference on Computer Vision and
  Pattern Recognition}, pages 6792--6802, 2022.

\bibitem{damen2018scaling}
Dima Damen, Hazel Doughty, Giovanni~Maria Farinella, Sanja Fidler, Antonino
  Furnari, Evangelos Kazakos, Davide Moltisanti, Jonathan Munro, Toby Perrett,
  Will Price, et~al.
\newblock Scaling egocentric vision: The epic-kitchens dataset.
\newblock In {\em Proceedings of the European Conference on Computer Vision
  (ECCV)}, pages 720--736, 2018.

\bibitem{das2018embodied}
Abhishek Das, Samyak Datta, Georgia Gkioxari, Stefan Lee, Devi Parikh, and
  Dhruv Batra.
\newblock Embodied question answering.
\newblock In {\em Proceedings of the IEEE Conference on Computer Vision and
  Pattern Recognition}, pages 1--10, 2018.

\bibitem{datta2022episodic}
Samyak Datta, Sameer Dharur, Vincent Cartillier, Ruta Desai, Mukul Khanna,
  Dhruv Batra, and Devi Parikh.
\newblock Episodic memory question answering.
\newblock In {\em Proceedings of the IEEE/CVF Conference on Computer Vision and
  Pattern Recognition}, pages 19119--19128, 2022.

\bibitem{deng2009imagenet}
Jia Deng, Wei Dong, Richard Socher, Li-Jia Li, Kai Li, and Li Fei-Fei.
\newblock Imagenet: A large-scale hierarchical image database.
\newblock In {\em Computer Vision and Pattern Recognition, 2009. CVPR 2009.
  IEEE Conference on}, pages 248--255. IEEE, 2009.

\bibitem{everingham2009pascal}
Mark Everingham, Luc Van~Gool, Christopher~KI Williams, John Winn, and Andrew
  Zisserman.
\newblock The pascal visual object classes (voc) challenge.
\newblock {\em International journal of computer vision}, 88:303--308, 2009.

\bibitem{fan2020few}
Qi Fan, Wei Zhuo, Chi-Keung Tang, and Yu-Wing Tai.
\newblock Few-shot object detection with attention-rpn and multi-relation
  detector.
\newblock In {\em Proceedings of the IEEE/CVF conference on computer vision and
  pattern recognition}, pages 4013--4022, 2020.

\bibitem{Geiger2012CVPR}
Andreas Geiger, Philip Lenz, and Raquel Urtasun.
\newblock Are we ready for autonomous driving? the kitti vision benchmark
  suite.
\newblock In {\em Conference on Computer Vision and Pattern Recognition
  (CVPR)}, 2012.

\bibitem{girshick2015fast}
Ross Girshick.
\newblock Fast r-cnn.
\newblock In {\em Proceedings of the IEEE international conference on computer
  vision}, pages 1440--1448, 2015.

\bibitem{girshick2014rich}
Ross Girshick, Jeff Donahue, Trevor Darrell, and Jitendra Malik.
\newblock Rich feature hierarchies for accurate object detection and semantic
  segmentation.
\newblock In {\em Proceedings of the IEEE conference on computer vision and
  pattern recognition}, pages 580--587, 2014.

\bibitem{grauman2022ego4d}
Kristen Grauman, Andrew Westbury, Eugene Byrne, Zachary Chavis, Antonino
  Furnari, Rohit Girdhar, Jackson Hamburger, Hao Jiang, Miao Liu, Xingyu Liu,
  et~al.
\newblock Ego4d: Around the world in 3,000 hours of egocentric video.
\newblock In {\em Proceedings of the IEEE/CVF Conference on Computer Vision and
  Pattern Recognition}, pages 18995--19012, 2022.

\bibitem{gupta2019lvis}
Agrim Gupta, Piotr Dollar, and Ross Girshick.
\newblock Lvis: A dataset for large vocabulary instance segmentation.
\newblock In {\em Proceedings of the IEEE/CVF conference on computer vision and
  pattern recognition}, pages 5356--5364, 2019.

\bibitem{guzov2021human}
Vladimir Guzov, Aymen Mir, Torsten Sattler, and Gerard Pons-Moll.
\newblock Human poseitioning system (hps): 3d human pose estimation and
  self-localization in large scenes from body-mounted sensors.
\newblock In {\em Proceedings of the IEEE/CVF Conference on Computer Vision and
  Pattern Recognition}, pages 4318--4329, 2021.

\bibitem{sparf}
Abdullah Hamdi, Bernard Ghanem, and Matthias Nie{\ss}ner.
\newblock Sparf: Large-scale learning of 3d sparse radiance fields from few
  input images.
\newblock {\em arXiv preprint arXiv:2212.09100}, 2022.

\bibitem{mvtn}
Abdullah Hamdi, Silvio Giancola, and Bernard Ghanem.
\newblock Mvtn: Multi-view transformation network for 3d shape recognition.
\newblock In {\em Proceedings of the IEEE/CVF International Conference on
  Computer Vision (ICCV)}, pages 1--11, October 2021.

\bibitem{hamdi2023voint}
Abdullah Hamdi, Silvio Giancola, and Bernard Ghanem.
\newblock Voint cloud: Multi-view point cloud representation for 3d
  understanding.
\newblock In {\em The Eleventh International Conference on Learning
  Representations}, 2023.

\bibitem{hartley2003multiple}
Richard Hartley and Andrew Zisserman.
\newblock {\em Multiple view geometry in computer vision}.
\newblock Cambridge university press, 2003.

\bibitem{he2017mask}
Kaiming He, Georgia Gkioxari, Piotr Doll{\'a}r, and Ross Girshick.
\newblock Mask r-cnn.
\newblock In {\em Proceedings of the IEEE international conference on computer
  vision}, pages 2961--2969, 2017.

\bibitem{he2015spatial}
Kaiming He, Xiangyu Zhang, Shaoqing Ren, and Jian Sun.
\newblock Spatial pyramid pooling in deep convolutional networks for visual
  recognition.
\newblock {\em IEEE transactions on pattern analysis and machine intelligence},
  37(9):1904--1916, 2015.

\bibitem{vars}
Jan Held, Anthony Cioppa, Silvio Giancola, Abdullah Hamdi, Bernard Ghanem, and
  Marc Van~Droogenbroeck.
\newblock Vars: Video assistant referee system for automated soccer decision
  making from multiple views.
\newblock In {\em Proceedings of the IEEE/CVF Conference on Computer Vision and
  Pattern Recognition}, pages 5085--5096, 2023.

\bibitem{huang2018apolloscape}
Xinyu Huang, Xinjing Cheng, Qichuan Geng, Binbin Cao, Dingfu Zhou, Peng Wang,
  Yuanqing Lin, and Ruigang Yang.
\newblock The apolloscape dataset for autonomous driving.
\newblock In {\em Proceedings of the IEEE conference on computer vision and
  pattern recognition workshops}, pages 954--960, 2018.

\bibitem{jiang2017seeing}
Hao Jiang and Kristen Grauman.
\newblock Seeing invisible poses: Estimating 3d body pose from egocentric
  video.
\newblock In {\em 2017 IEEE Conference on Computer Vision and Pattern
  Recognition (CVPR)}, pages 3501--3509. IEEE, 2017.

\bibitem{joseph2021towards}
KJ Joseph, Salman Khan, Fahad~Shahbaz Khan, and Vineeth~N Balasubramanian.
\newblock Towards open world object detection.
\newblock In {\em Proceedings of the IEEE/CVF Conference on Computer Vision and
  Pattern Recognition}, pages 5830--5840, 2021.

\bibitem{kang2019few}
Bingyi Kang, Zhuang Liu, Xin Wang, Fisher Yu, Jiashi Feng, and Trevor Darrell.
\newblock Few-shot object detection via feature reweighting.
\newblock In {\em Proceedings of the IEEE/CVF International Conference on
  Computer Vision}, pages 8420--8429, 2019.

\bibitem{karpathy2014large}
Andrej Karpathy, George Toderici, Sanketh Shetty, Thomas Leung, Rahul
  Sukthankar, and Li Fei-Fei.
\newblock Large-scale video classification with convolutional neural networks.
\newblock In {\em Proceedings of the IEEE conference on Computer Vision and
  Pattern Recognition}, pages 1725--1732, 2014.

\bibitem{kazakos2021little}
Evangelos Kazakos, Jaesung Huh, Arsha Nagrani, Andrew Zisserman, and Dima
  Damen.
\newblock With a little help from my temporal context: Multimodal egocentric
  action recognition.
\newblock {\em arXiv preprint arXiv:2111.01024}, 2021.

\bibitem{kazakos2019epic}
Evangelos Kazakos, Arsha Nagrani, Andrew Zisserman, and Dima Damen.
\newblock Epic-fusion: Audio-visual temporal binding for egocentric action
  recognition.
\newblock In {\em Proceedings of the IEEE/CVF International Conference on
  Computer Vision}, pages 5492--5501, 2019.

\bibitem{kendall2015posenet}
Alex Kendall, Matthew Grimes, and Roberto Cipolla.
\newblock Posenet: A convolutional network for real-time 6-dof camera
  relocalization.
\newblock In {\em Proceedings of the IEEE international conference on computer
  vision}, pages 2938--2946, 2015.

\bibitem{kuznetsova2020open}
Alina Kuznetsova, Hassan Rom, Neil Alldrin, Jasper Uijlings, Ivan Krasin, Jordi
  Pont-Tuset, Shahab Kamali, Stefan Popov, Matteo Malloci, Alexander
  Kolesnikov, et~al.
\newblock The open images dataset v4: Unified image classification, object
  detection, and visual relationship detection at scale.
\newblock {\em International Journal of Computer Vision}, 128(7):1956--1981,
  2020.

\bibitem{labbe2019rtab}
Mathieu Labb{\'e} and Fran{\c{c}}ois Michaud.
\newblock Rtab-map as an open-source lidar and visual simultaneous localization
  and mapping library for large-scale and long-term online operation.
\newblock {\em Journal of Field Robotics}, 36(2):416--446, 2019.

\bibitem{Li2022sctn}
Bing Li, Cheng Zheng, Silvio Giancola, and Bernard Ghanem.
\newblock {SCTN}: Sparse convolution-transformer network for scene flow
  estimation.
\newblock In {\em AAAI}, 2022.

\bibitem{li2023ego}
Jiaman Li, Karen Liu, and Jiajun Wu.
\newblock Ego-body pose estimation via ego-head pose estimation.
\newblock In {\em Proceedings of the IEEE/CVF Conference on Computer Vision and
  Pattern Recognition}, pages 17142--17151, 2023.

\bibitem{li2018eye}
Yin Li, Miao Liu, and James~M Rehg.
\newblock In the eye of beholder: Joint learning of gaze and actions in first
  person video.
\newblock In {\em Proceedings of the European conference on computer vision
  (ECCV)}, pages 619--635, 2018.

\bibitem{li2021ego}
Yanghao Li, Tushar Nagarajan, Bo Xiong, and Kristen Grauman.
\newblock Ego-exo: Transferring visual representations from third-person to
  first-person videos.
\newblock In {\em Proceedings of the IEEE/CVF Conference on Computer Vision and
  Pattern Recognition}, pages 6943--6953, 2021.

\bibitem{lin2017feature}
Tsung-Yi Lin, Piotr Doll{\'a}r, Ross Girshick, Kaiming He, Bharath Hariharan,
  and Serge Belongie.
\newblock Feature pyramid networks for object detection.
\newblock In {\em Proceedings of the IEEE conference on computer vision and
  pattern recognition}, pages 2117--2125, 2017.

\bibitem{lin2017focal}
Tsung-Yi Lin, Priya Goyal, Ross Girshick, Kaiming He, and Piotr Doll{\'a}r.
\newblock Focal loss for dense object detection.
\newblock In {\em Proceedings of the IEEE international conference on computer
  vision}, pages 2980--2988, 2017.

\bibitem{lin2014microsoft}
Tsung-Yi Lin, Michael Maire, Serge Belongie, James Hays, Pietro Perona, Deva
  Ramanan, Piotr Doll{\'a}r, and C~Lawrence Zitnick.
\newblock Microsoft coco: Common objects in context.
\newblock In {\em Computer Vision--ECCV 2014: 13th European Conference, Zurich,
  Switzerland, September 6-12, 2014, Proceedings, Part V 13}, pages 740--755.
  Springer, 2014.

\bibitem{coco}
Tsung-Yi Lin, Michael Maire, Serge Belongie, James Hays, Pietro Perona, Deva
  Ramanan, Piotr Dollár, and C.~Lawrence Zitnick.
\newblock Microsoft coco: Common objects in context.
\newblock In {\em European Conference on Computer Vision (ECCV)}, Zürich,
  2014.
\newblock Oral.

\bibitem{liu2022depth}
Sheng Liu, Xiaohan Nie, and Raffay Hamid.
\newblock Depth-guided sparse structure-from-motion for movies and tv shows.
\newblock In {\em Proceedings of the IEEE/CVF Conference on Computer Vision and
  Pattern Recognition}, pages 15980--15989, 2022.

\bibitem{liu20224d}
Yunze Liu, Yun Liu, Che Jiang, Z Fu, K Lyu, W Wan, H Shen, B Liang, He Wang,
  and Li~Yi Hoi4d.
\newblock A 4d egocentric dataset for category-level human-object interaction.
\newblock In {\em IEEE/CVF Conf. Comput. Vis. Pattern Recog.(CVPR), 2022b},
  volume~1, 2022.

\bibitem{luo2021m3dssd}
Shujie Luo, Hang Dai, Ling Shao, and Yong Ding.
\newblock M3dssd: Monocular 3d single stage object detector.
\newblock In {\em Proceedings of the IEEE/CVF Conference on Computer Vision and
  Pattern Recognition}, pages 6145--6154, 2021.

\bibitem{munro2020multi}
Jonathan Munro and Dima Damen.
\newblock Multi-modal domain adaptation for fine-grained action recognition.
\newblock In {\em Proceedings of the IEEE/CVF conference on computer vision and
  pattern recognition}, pages 122--132, 2020.

\bibitem{mur2015orb}
Raul Mur-Artal, Jose Maria~Martinez Montiel, and Juan~D Tardos.
\newblock Orb-slam: a versatile and accurate monocular slam system.
\newblock {\em IEEE transactions on robotics}, 31(5):1147--1163, 2015.

\bibitem{nagarajan2022egocentric}
Tushar Nagarajan, Santhosh~Kumar Ramakrishnan, Ruta Desai, James Hillis, and
  Kristen Grauman.
\newblock Egocentric scene context for human-centric environment understanding
  from video.
\newblock {\em arXiv preprint arXiv:2207.11365}, 2022.

\bibitem{Silberman:ECCV12}
Pushmeet~Kohli Nathan~Silberman, Derek~Hoiem and Rob Fergus.
\newblock Indoor segmentation and support inference from rgbd images.
\newblock In {\em ECCV}, 2012.

\bibitem{nguyen2016recognition}
Thi-Hoa-Cuc Nguyen, Jean-Christophe Nebel, and Francisco Florez-Revuelta.
\newblock Recognition of activities of daily living with egocentric vision: A
  review.
\newblock {\em Sensors}, 16(1):72, 2016.

\bibitem{ozyecsil2017survey}
Onur {\"O}zye{\c{s}}il, Vladislav Voroninski, Ronen Basri, and Amit Singer.
\newblock A survey of structure from motion*.
\newblock {\em Acta Numerica}, 26:305--364, 2017.

\bibitem{patra2019ego}
Suvam Patra, Kartikeya Gupta, Faran Ahmad, Chetan Arora, and Subhashis
  Banerjee.
\newblock Ego-slam: A robust monocular slam for egocentric videos.
\newblock In {\em 2019 IEEE Winter Conference on Applications of Computer
  Vision (WACV)}, pages 31--40. IEEE, 2019.

\bibitem{perez2020incremental}
Juan-Manuel Perez-Rua, Xiatian Zhu, Timothy~M Hospedales, and Tao Xiang.
\newblock Incremental few-shot object detection.
\newblock In {\em Proceedings of the IEEE/CVF Conference on Computer Vision and
  Pattern Recognition}, pages 13846--13855, 2020.

\bibitem{pix4point}
Guocheng Qian, Xingdi Zhang, Abdullah Hamdi, and Bernard Ghanem.
\newblock Pix4point: Image pretrained transformers for 3d point cloud
  understanding.
\newblock {\em arXiv preprint arXiv:2208.12259}, 2022.

\bibitem{ramazanova2023owl}
Merey Ramazanova, Victor Escorcia, Fabian~Caba Heilbron, Chen Zhao, and Bernard
  Ghanem.
\newblock Owl (observe, watch, listen): Localizing actions in egocentric video
  via audiovisual temporal context.
\newblock In {\em Proceedings of the IEEE/CVF Conference on Computer Vision and
  Pattern Recognition Workshop (CVPRW)}, 2023.

\bibitem{ranftl2021vision}
Ren{\'e} Ranftl, Alexey Bochkovskiy, and Vladlen Koltun.
\newblock Vision transformers for dense prediction.
\newblock In {\em Proceedings of the IEEE/CVF International Conference on
  Computer Vision}, pages 12179--12188, 2021.

\bibitem{redmon2016you}
Joseph Redmon, Santosh Divvala, Ross Girshick, and Ali Farhadi.
\newblock You only look once: Unified, real-time object detection.
\newblock In {\em Proceedings of the IEEE conference on computer vision and
  pattern recognition}, pages 779--788, 2016.

\bibitem{faster-rcnn}
Shaoqing Ren, Kaiming He, Ross Girshick, and Jian Sun.
\newblock Faster r-cnn: Towards real-time object detection with region proposal
  networks.
\newblock In C. Cortes, N.~D. Lawrence, D.~D. Lee, M. Sugiyama, and R. Garnett,
  editors, {\em Advances in Neural Information Processing Systems 28}, pages
  91--99. Curran Associates, Inc., 2015.

\bibitem{ren2015faster}
Shaoqing Ren, Kaiming He, Ross Girshick, and Jian Sun.
\newblock Faster r-cnn: Towards real-time object detection with region proposal
  networks.
\newblock {\em Advances in neural information processing systems}, 28, 2015.

\bibitem{rhodin2016egocap}
Helge Rhodin, Christian Richardt, Dan Casas, Eldar Insafutdinov, Mohammad
  Shafiei, Hans-Peter Seidel, Bernt Schiele, and Christian Theobalt.
\newblock Egocap: egocentric marker-less motion capture with two fisheye
  cameras.
\newblock {\em ACM Transactions on Graphics (TOG)}, 35(6):1--11, 2016.

\bibitem{rosano2021embodied}
Marco Rosano, Antonino Furnari, Luigi Gulino, and Giovanni~Maria Farinella.
\newblock On embodied visual navigation in real environments through habitat.
\newblock In {\em 2020 25th International Conference on Pattern Recognition
  (ICPR)}, pages 9740--9747. IEEE, 2021.

\bibitem{rukhovich2022imvoxelnet}
Danila Rukhovich, Anna Vorontsova, and Anton Konushin.
\newblock Imvoxelnet: Image to voxels projection for monocular and multi-view
  general-purpose 3d object detection.
\newblock In {\em Proceedings of the IEEE/CVF Winter Conference on Applications
  of Computer Vision}, pages 2397--2406, 2022.

\bibitem{schoenberger2016sfm}
Johannes~Lutz Sch\"{o}nberger and Jan-Michael Frahm.
\newblock Structure-from-motion revisited.
\newblock In {\em Conference on Computer Vision and Pattern Recognition
  (CVPR)}, 2016.

\bibitem{scipy_peak}
{SciPy Contributors}.
\newblock {SciPy}: Scientific library for {Python} - scipy.signal.find\_peaks.
\newblock Online, 2021.

\bibitem{sener2020temporal}
Fadime Sener, Dipika Singhania, and Angela Yao.
\newblock Temporal aggregate representations for long-range video
  understanding.
\newblock In {\em Computer Vision--ECCV 2020: 16th European Conference,
  Glasgow, UK, August 23--28, 2020, Proceedings, Part XVI 16}, pages 154--171.
  Springer, 2020.

\bibitem{sigurdsson2018charades}
Gunnar~A Sigurdsson, Abhinav Gupta, Cordelia Schmid, Ali Farhadi, and Karteek
  Alahari.
\newblock Charades-ego: A large-scale dataset of paired third and first person
  videos.
\newblock {\em arXiv preprint arXiv:1804.09626}, 2018.

\bibitem{soldan2022mad}
Mattia Soldan, Alejandro Pardo, Juan~Le{\'o}n Alc{\'a}zar, Fabian Caba, Chen
  Zhao, Silvio Giancola, and Bernard Ghanem.
\newblock Mad: A scalable dataset for language grounding in videos from movie
  audio descriptions.
\newblock In {\em Proceedings of the IEEE/CVF Conference on Computer Vision and
  Pattern Recognition}, pages 5026--5035, 2022.

\bibitem{song2015sun}
Shuran Song, Samuel~P Lichtenberg, and Jianxiong Xiao.
\newblock Sun rgb-d: A rgb-d scene understanding benchmark suite.
\newblock In {\em Proceedings of the IEEE conference on computer vision and
  pattern recognition}, pages 567--576, 2015.

\bibitem{teed2018deepv2d}
Zachary Teed and Jia Deng.
\newblock Deepv2d: Video to depth with differentiable structure from motion.
\newblock {\em arXiv preprint arXiv:1812.04605}, 2018.

\bibitem{teed2021droid}
Zachary Teed and Jia Deng.
\newblock Droid-slam: Deep visual slam for monocular, stereo, and rgb-d
  cameras.
\newblock {\em Advances in neural information processing systems},
  34:16558--16569, 2021.

\bibitem{thapar2021anonymizing}
Daksh Thapar, Aditya Nigam, and Chetan Arora.
\newblock Anonymizing egocentric videos.
\newblock In {\em Proceedings of the IEEE/CVF International Conference on
  Computer Vision}, pages 2320--2329, 2021.

\bibitem{tian2019fcos}
Zhi Tian, Chunhua Shen, Hao Chen, and Tong He.
\newblock Fcos: Fully convolutional one-stage object detection.
\newblock In {\em Proceedings of the IEEE/CVF international conference on
  computer vision}, pages 9627--9636, 2019.

\bibitem{tome2020selfpose}
Denis Tome, Thiemo Alldieck, Patrick Peluse, Gerard Pons-Moll, Lourdes Agapito,
  Hernan Badino, and Fernando De~la Torre.
\newblock Selfpose: 3d egocentric pose estimation from a headset mounted
  camera.
\newblock {\em arXiv preprint arXiv:2011.01519}, 2020.

\bibitem{tome2019xr}
Denis Tome, Patrick Peluse, Lourdes Agapito, and Hernan Badino.
\newblock xr-egopose: Egocentric 3d human pose from an hmd camera.
\newblock In {\em Proceedings of the IEEE/CVF International Conference on
  Computer Vision}, pages 7728--7738, 2019.

\bibitem{tschernezki2022neural}
Vadim Tschernezki, Iro Laina, Diane Larlus, and Andrea Vedaldi.
\newblock Neural feature fusion fields: 3d distillation of self-supervised 2d
  image representations.
\newblock {\em arXiv preprint arXiv:2209.03494}, 2022.

\bibitem{tschernezki2021neuraldiff}
Vadim Tschernezki, Diane Larlus, and Andrea Vedaldi.
\newblock Neuraldiff: Segmenting 3d objects that move in egocentric videos.
\newblock In {\em 2021 International Conference on 3D Vision (3DV)}, pages
  910--919. IEEE, 2021.

\bibitem{tulving2002episodic}
Endel Tulving.
\newblock Episodic memory: From mind to brain.
\newblock {\em Annual review of psychology}, 53(1):1--25, 2002.

\bibitem{vijayanarasimhan2017sfm}
Sudheendra Vijayanarasimhan, Susanna Ricco, Cordelia Schmid, Rahul Sukthankar,
  and Katerina Fragkiadaki.
\newblock Sfm-net: Learning of structure and motion from video.
\newblock {\em arXiv preprint arXiv:1704.07804}, 2017.

\bibitem{wang2021estimating}
Jian Wang, Lingjie Liu, Weipeng Xu, Kripasindhu Sarkar, and Christian Theobalt.
\newblock Estimating egocentric 3d human pose in global space.
\newblock In {\em Proceedings of the IEEE/CVF International Conference on
  Computer Vision}, pages 11500--11509, 2021.

\bibitem{wu2019long}
Chao-Yuan Wu, Christoph Feichtenhofer, Haoqi Fan, Kaiming He, Philipp
  Krahenbuhl, and Ross Girshick.
\newblock Long-term feature banks for detailed video understanding.
\newblock In {\em Proceedings of the IEEE/CVF Conference on Computer Vision and
  Pattern Recognition}, pages 284--293, 2019.

\bibitem{xu2022negative}
Mengmeng Xu, Cheng-Yang Fu, Yanghao Li, Bernard Ghanem, Juan-Manuel Perez-Rua,
  and Tao Xiang.
\newblock Negative frames matter in egocentric visual query 2d localization.
\newblock {\em arXiv preprint arXiv:2208.01949}, 2022.

\bibitem{xu2022my}
Mengmeng Xu, Yanghao Li, Cheng-Yang Fu, Bernard Ghanem, Tao Xiang, and
  Juan-Manuel Perez-Rua.
\newblock Where is my wallet? modeling object proposal sets for egocentric
  visual query localization.
\newblock {\em arXiv preprint arXiv:2211.10528}, 2022.

\bibitem{xu2020g}
Mengmeng Xu, Chen Zhao, David~S Rojas, Ali Thabet, and Bernard Ghanem.
\newblock G-tad: Sub-graph localization for temporal action detection.
\newblock In {\em Proceedings of the IEEE/CVF Conference on Computer Vision and
  Pattern Recognition}, 2020.

\bibitem{xu2019mo}
Weipeng Xu, Avishek Chatterjee, Michael Zollhoefer, Helge Rhodin, Pascal Fua,
  Hans-Peter Seidel, and Christian Theobalt.
\newblock Mo 2 cap 2: Real-time mobile 3d motion capture with a cap-mounted
  fisheye camera.
\newblock {\em IEEE transactions on visualization and computer graphics},
  25(5):2093--2101, 2019.

\bibitem{yin2022sylph}
Li Yin, Juan~M Perez-Rua, and Kevin~J Liang.
\newblock Sylph: A hypernetwork framework for incremental few-shot object
  detection.
\newblock In {\em Proceedings of the IEEE/CVF Conference on Computer Vision and
  Pattern Recognition}, pages 9035--9045, 2022.

\bibitem{zhang2022egobody}
Siwei Zhang, Qianli Ma, Yan Zhang, Zhiyin Qian, Taein Kwon, Marc Pollefeys,
  Federica Bogo, and Siyu Tang.
\newblock Egobody: Human body shape and motion of interacting people from
  head-mounted devices.
\newblock In {\em European Conference on Computer Vision}, pages 180--200.
  Springer, 2022.

\bibitem{zhang2022structure}
Zhoutong Zhang, Forrester Cole, Zhengqi Li, Michael Rubinstein, Noah Snavely,
  and William~T Freeman.
\newblock Structure and motion from casual videos.
\newblock In {\em Computer Vision--ECCV 2022: 17th European Conference, Tel
  Aviv, Israel, October 23--27, 2022, Proceedings, Part XXXIII}, pages 20--37.
  Springer, 2022.

\bibitem{zhao2023re2tal}
Chen Zhao, Shuming Liu, Karttikeya Mangalam, and Bernard Ghanem.
\newblock {Re2TAL}: Rewiring pretrained video backbones for reversible temporal
  action localization.
\newblock In {\em Proceedings of the IEEE/CVF Conference on Computer Vision and
  Pattern Recognition (CVPR)}, 2023.

\bibitem{zhao2021video}
Chen Zhao, Ali~K Thabet, and Bernard Ghanem.
\newblock Video self-stitching graph network for temporal action localization.
\newblock In {\em Proceedings of the IEEE/CVF International Conference on
  Computer Vision (ICCV)}, 2021.

\bibitem{zhao2022particlesfm}
Wang Zhao, Shaohui Liu, Hengkai Guo, Wenping Wang, and Yong-Jin Liu.
\newblock Particlesfm: Exploiting dense point trajectories for localizing
  moving cameras in the wild.
\newblock In {\em Computer Vision--ECCV 2022: 17th European Conference, Tel
  Aviv, Israel, October 23--27, 2022, Proceedings, Part XXXII}, pages 523--542.
  Springer, 2022.

\bibitem{zhao2022revisiting}
Xiaowei Zhao, Xianglong Liu, Yifan Shen, Yuqing Ma, Yixuan Qiao, and Duorui
  Wang.
\newblock Revisiting open world object detection.
\newblock {\em arXiv preprint arXiv:2201.00471}, 2022.

\bibitem{zhao2019object}
Zhong-Qiu Zhao, Peng Zheng, Shou-tao Xu, and Xindong Wu.
\newblock Object detection with deep learning: A review.
\newblock {\em IEEE transactions on neural networks and learning systems},
  30(11):3212--3232, 2019.

\bibitem{zhou2017unsupervised}
Tinghui Zhou, Matthew Brown, Noah Snavely, and David~G Lowe.
\newblock Unsupervised learning of depth and ego-motion from video.
\newblock In {\em Proceedings of the IEEE conference on computer vision and
  pattern recognition}, pages 1851--1858, 2017.

\end{thebibliography}
}

\thispagestyle{empty}
\appendix

\section{Reproducibility}\label{sec:repro}
\textbf{The code and models are released at \url{https://github.com/Wayne-Mai/EgoLoc}.}

\subsection{Structure-from-Motion}
In this section, we provide the detailed hyperparameters we used for
the first stage, Structure from Motion (SfM).
\subsubsection{Camera Intrinsics}
To be robust to motion blur, we subsample 100 contiguous non-blurry frames from the videos selected by the variance of Laplacian greater than 100, which will be fed into COLMAP~\cite{schoenberger2016sfm} auto reconstruction to estimate camera intrinsic at first.
To model the fisheye distortion in egocentric videos, we choose the \textit{RADIAL\_FISHEYE} camera.
Then we use COLMAP \textit{automatic\_reconstructor} and enforce a single camera model to estimate the camera intrinsics from a dense reconstruction.

\subsubsection{Camera Poses}
We propose to use sparse mapping from COLMAP to estimate the camera poses among the video frames directly.
In this way, we no longer need to worry about the inconsistency between scanned frames and video frames, but we also get rid of the strong assumption that there must be an existing 3D scan of the video scene.
This saves a lot of computing power and resources for scan creation and makes our method easy to migrate and work in real-world applications.

Given an egocentric video $\mathcal{V}$ with frames $\{k_i,i\in \mathcal{I}\}$ where $\mathcal{I}$ is the total number of frames, we first extract the features of the frames and match them among the video frames.
Since the video frames are ordered, we adopt the sequential matching functionality from COLMAP, which leads to a fast SfM with acceptable quality for the thousands of images to structure.
For each frame $k_i$, COLMAP performs a coarse matching with its temporal neighboring frames $\{k_{i-\frac{w}{2}},...,k_{i+\frac{w}{2}}\}$ in a window size of $w$.
It creates a scene graph that matches temporally contiguous images.
We feed our video frames with the scene graphs into our COLMAP solver, which reconstructs a sparse 3D map $\mathcal{M}$ from the video keyframes $\{k_{m_0},k_{m_1},...\}$ that were successfully registered.

The map $\mathcal{M}$ with sparse feature points and pose images will be used to extract the camera poses we are interested in, i.e., the peak response frames that contain the query object, later on.
Note that in some cases, not all video frames can be registered into the map $\mathcal{M}$ because of the extreme dynamics of egocentric videos.
Therefore, in some cases, we don't have camera poses for query frames or response frames, resulting in a failed instance of \textit{Query with Pose} (QwP).

\subsubsection{COLMAP Hyperparameters}
We empirically explore the proper hyperparameters for egocentric videos and present the detailed hyperparameters we used in Table~\ref{tb:sup_colmap}.
\begin{table}[!htb]
\centering
\resizebox{0.5\textwidth}{!}{%
\begin{tabular}{l|l}
\hline
Feature\_extractor                      & \textit{SIFT}                                 \\ \hline
Num Camera                              & \textit{1}                                    \\ \hline
Matcher                                 & \textit{sequential matcher}                   \\ \hline
Matcher.vocab\_tree                     & \textit{vocab\_tree\_flickr100K\_words1M.bin} \\ \hline
Matcher.window\_size                     & \textit{10} \\ \hline
Mapper.abs\_pose\_min\_num\_inliers     & \textit{15}                                   \\ \hline
Mapper.init\_min\_tri\_angle            & \textit{12}                                   \\ \hline
Mapper.abs\_pose\_min\_inlier\_ratio    & \textit{0.2}                                  \\ \hline
Mapper.ba\_global\_max\_num\_iterations & \textit{30}                                   \\ \hline
\end{tabular}%
}
\caption{
        \textbf{Hyperparameters for COLMAP.} 
    }
    \label{tb:sup_colmap}
\end{table}

\subsection{Robustness to depth estimation} Since we don't have direct access to GT depth for video frames, we add noise to DPT~\cite{ranftl2021vision} to test the robustness. We consider two common errors in depth estimation: scale error $k$ and shift error $b$. Then the depth prediction $d$ from DPT becomes $\hat{d}=kd+b$.
As shown in Table~\ref{tb:depth}, a small Gaussian noise will give a mild drop in performance, while a completely random depth will break the whole prediction.

\begin{table}[!hbt]
\centering
\resizebox{0.99\linewidth}{!}{%
\begin{tabular}{l|ccccc}
\toprule
\textbf{Depth Estimator (Validation Set)} & \textbf{Succ\%↑} & \textbf{Succ*\%↑} & \textbf{L2↓} & \textbf{Angle↓} & \textbf{QwP\%↑} \\ \midrule
DPT $\{k=1,b=0\}$                 & \textbf{80.49} & \textbf{98.14} & \textbf{1.45} & \textbf{0.61} & \textbf{82.32} \\ 
Noise $\{k\sim \mathcal{N}(1,0.2),b\sim \mathcal{N}(0,0.2)\}$ & 78.66 & 95.65 & 2.41 & 0.67 & 82.32 \\
Random $\{k=0,b\sim \mathcal{U}(0.1,10)\}$ & 34.15 & 43.42 & 5.77 & 0.89 & 82.32    \\
\bottomrule
\end{tabular}%
}
\vspace{2pt}
\caption{
        \textbf{Depth Robustness}
        (\textit{Noise}): adding Gaussian noise to the depth scale and shift on DPT prediction,  (\textit{Random}): uniformly sampled from $0.1\sim 10$ meters.}
    \label{tb:depth}
\end{table}

\subsection{VQ2D Detection}
We follow most of the settings from Ego4D~\cite{grauman2022ego4d} VQ2D baseline for our detector without major revision. The 2D detector backbone, i.e., the feature extractor $\mathcal{F}$, is adopted from Faster-RCNN~\cite{faster-rcnn}. We refer our readers to Faster-RCNN and Ego4D Episodic Memory VQ2D Benchmark\footnote{\url{https://github.com/EGO4D/episodic-memory/tree/main/VQ2D}} for more details.

\subsubsection{Siam-RCNN}
The Siam-RCNN is the detector architecture we adopt for object detection from videos.

To detect the query object in a video frame $k_i$, a pre-trained Region Proposal Network (RPN)~\cite{faster-rcnn} with a Feature Pyramid Network (FPN)~\cite{lin2017feature} backbone is utilized to generate a set of bounding box proposals $\{b_{i_0},b_{i_1},...\}$. These proposals are then processed through the RoI-Align operation~\cite{he2017mask} to extract visual features for each box $\{ \mathcal{F}(b_{i_0}),\mathcal{F}(b_{i_1}),...\}$. Meanwhile, we also extract features $ \mathcal{F}(v)$ for the visual crop $v$ using the same FPN backbone. 

The SiamHead $\mathcal{S}$ compares $\{ \mathcal{F}(b_0),\mathcal{F}(b_1),...\}$  with $ \mathcal{F}(v)$. The SiamHead $\mathcal{S}$ consists of a convolutional projection module $\mathcal{P}$ to project those features into 1024-D, and a bilinear operation layer with sigmoid activation $\sigma$ to output a similarity score between $[0,1]$:

\begin{align}
    s_i=\sigma( \mathcal{P}(\mathcal{F}(b_i))^T W \mathcal{P}(\mathcal{F}(v)) + bias  )
    \label{eq:similarity},
\end{align}
for all the bounding box proposals: $\{s_{i_1},s_{i_2}...\}$.
Then we select the Top-1 bounding box proposal as $b_{i}$ for frame $k_i$. 

The projection module $\mathcal{P}$ in SiamHead $\mathcal{S}$ architecture comprises four residual blocks and is succeeded by average pooling. It also includes a two-layer multi-layer perceptron (MLP) that utilizes ReLU activation and has a hidden size of 1024-D.

\begin{figure}[t]
    \centering
    \includegraphics[width=\linewidth] {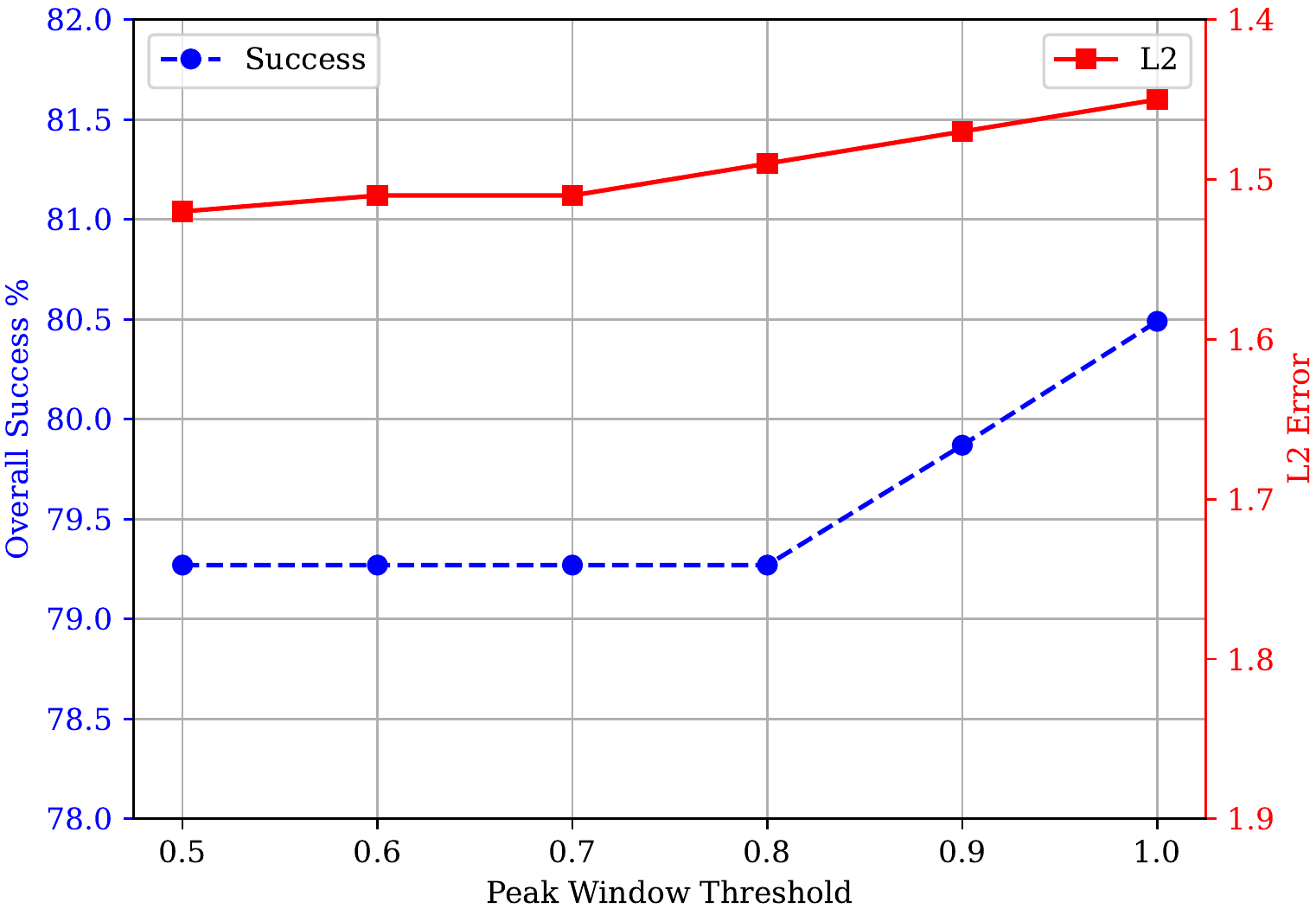}
    \caption{
        \textbf{Peak Window.} We ablate on the threshold for peak window selection and show performance metrics. Note that L2 Error axis is inverted to align the improvement direction of both L2 and success metrics. 
    }
    \label{fig:peak}
\end{figure}

\subsection{Peak Window} We also investigated the aggregation prediction by the neighboring frames of peak frames as objects usually appear in the video as a temporal window. Given a window threshold and peak frame $k_{p}$, we search and add those frames forward and backward, until the detection score doesn't satisfy: $s_{p\pm i} \geq window\_threshold * s_{p}$. We can find in Figure~\ref{fig:peak} that narrowing the peak window can gradually boost the result. The best result is when we select the peak responses only.

\subsection{Training}
For fair comparison and brevity, we use the same training split and subsampled video frames, which consist of positive frames $D_y$ and negative frames $D_n$ indicating the presence of the object or not, as Ego4D VQ2D baseline. 

The 2D Faster-RCNN~\cite{faster-rcnn} backbone is pretrained on MS-COCO~\cite{coco} and frozen. We only train the Siamese head $\mathcal{S}$ in VQ2D~\cite{grauman2022ego4d} train set. 
We define the cross entropy loss function $\mathcal{L}$ for $\mathcal{S}$ for the similarity score $s_{b,v}=\mathcal{S}\left(\mathcal{F}(b),\mathcal{F}(v)\right)$ between features from frame bbox top proposal $b$ and visual crop $v$ of query object $o$:

\begin{align}
    \mathcal{L}=-\frac{1}{|D_y \cup D_n|} (\sum_{y\in D_y}\log (s_{y,v})+ \sum_{n\in D_n}\log (1-s_{n,v}).
    \label{eq:loss}
\end{align}

Both positives $y$ and negatives $n$ are defined based on proposals generated by the RPN, with IoU threshold$=0.5$ to the groundtruth response bbox annotation.

After training for 300,000 iterations with an initial learning rate of 0.02, we apply a 0.1× decay after 200,000 iterations. Additionally, we extract backbone features from the \textit{p3} layer of FPN.

\begin{table}[!htb]
\centering
\begin{tabular}{l|l}
\hline
distance    & 25  \\ \hline
width       & 3   \\ \hline
prominence  & 0.2 \\ \hline
wlen        & 50  \\ \hline
rel\_height & 0.5 \\ \hline
\end{tabular}%
\caption{
        \textbf{Parameters for peak search.} 
    }
    \label{tb:sup_peak}
\end{table}
\subsubsection{Peak Selection}
We can get a tuple of $\{(k_{0},b_{0},s_{0}),...,(k_{q},b_{q},s_{q})\}$.
After smoothing the scores with a median filter of $window\_size=5$, we search for the response peaks and select our response frames accordingly.
We select the peaks that satisfy Table~\ref{tb:sup_peak} using~\cite{scipy_peak}. `Distance' is the required minimal horizontal distance in samples between neighboring peaks. `Width' is the required width of peaks in samples. `Prominence' is the required prominence of peaks. `Wlen' is used for the calculation of the peak prominences.

\subsection{Registration}
The poses from COLMAP are independent of Matterport world coordinate system. 
We need to align our poses with the Matterport scan because annotators make the ground truth annotations in those scans with respect to the Matterport scan coordinate system.
Also, since the poses from SfM have the \textit{scale ambiguity}~\cite{hartley2003multiple} issue, we align the COLMAP reconstruction to Matterport scan coordinate system as post-processing. 
To evaluate our EgoLoc performance, we render at least three images from the Matterport scan with known camera poses and then perform a $Sim3$ transformation to align COLMAP coordinate system.

We use the $model\_aligner$ function provided by COLMAP to do the alignment between these two coordinate systems. We render at least three images from the Matterport scan with known camera poses. Then we use the script in Lst.~\ref{lst:align} to estimate the \textit{Sim3} transform between our coordinate system reconstructed by COLMAP and Matterport Scan. In this way, we can evaluate our results in the same coordinate system where the annotators annotate the ground truth 3D bboxes.

\begin{lstlisting}[language=Python, caption={Registration},label={lst:align},]    
subprocess.run([
    'colmap', 'model_aligner', 
    '--ref_images_path','matterport_renderings',
    '--ref_is_gps', '0',
    '--robust_alignment', '1', '--alignment_type', 'custom',
    '--estimate_scale', '1', '--robust_alignment_max_error', '25'
])
\end{lstlisting}

\section{Supporting information}\label{sec:support}
\subsection{About GT VQ2D Annotation}
\textit{Groundtruth Track} stands for the groundtruth tracking result provided by Ego4D VQ2D annotations, a temporal consecutive frame set showing the object's last appearance.
The groundtruth response track is specified as $r$, which is a temporally contiguous set of bounding boxes surrounding the object $o$ in each frame:
$$r=\{r_s,r_{s+1},...,r_{e-1},r_e\},$$
where $s$ is the frame where the object $o$ (at least partially) enters the camera wearer's field of view, $e$ is the frame where the object exits the camera wearer's field of view, and $r_i$ is a bounding box $(x,y,w,h)$ in frame $i$. If the object appears multiple times in the video, the response only refers to the `most recent appearance' of the object in the past, i.e., the response track, which minimizes $q-r_e$ with $q>r_e$.

Therefore, we don't have groundtruth detection annotations for all the frames in Ego4D, so we can't evaluate our proposed 2D response detection in this work.

\begin{figure*}[t]
    \centering
    \includegraphics[width=0.99\linewidth] {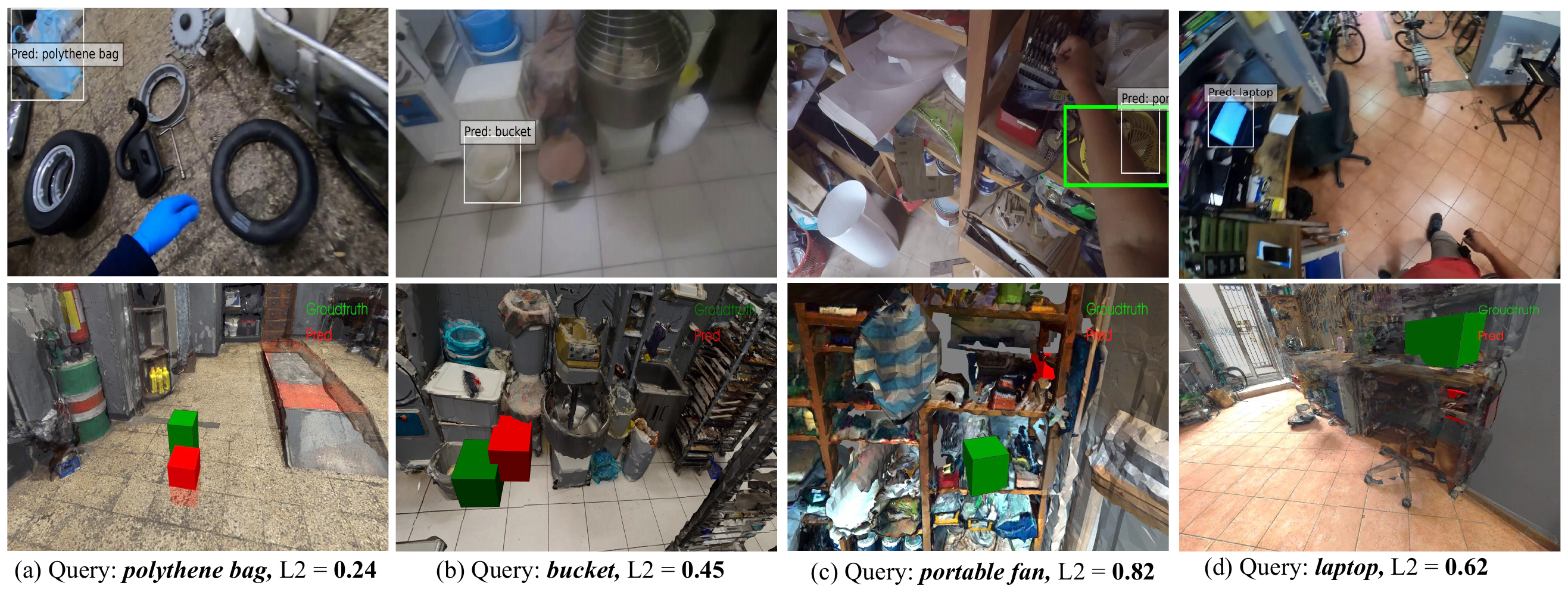}
    \caption{ \textbf{Visualization of 2D responses and 3D Localization.}
    }
    \label{fig:sup_vis_final}
\end{figure*}

\subsection{About Visualization}
We provide more visualization from our EgoLoc in the paper in Figure~\ref{fig:sup_vis_final}. 

\begin{enumerate}
    \item For 2D visualization, we use white to represent our predictions and green to represent the groundtruth.
    \item For 2D visualization, because we only have access to the groundtruth response track instead of groundtruth detection, we don't have groundtruth 2D bbox in some cases but only our own prediction bbox in white.
    \item For 3D visualization, note that we actually don't know the size and rotation of 3D bbox during prediction.
    But we use the size and rotation from groundtruth annotations during our visualization
    with our predicted 3D vector $(x,y,z)$ as the center of 3D bbox.
\end{enumerate}

\subsection{About Scenes}
\begin{figure*}[!htb]
    \centering
    \subfloat[Scooter mechanic]{\includegraphics[width=0.4\textwidth]{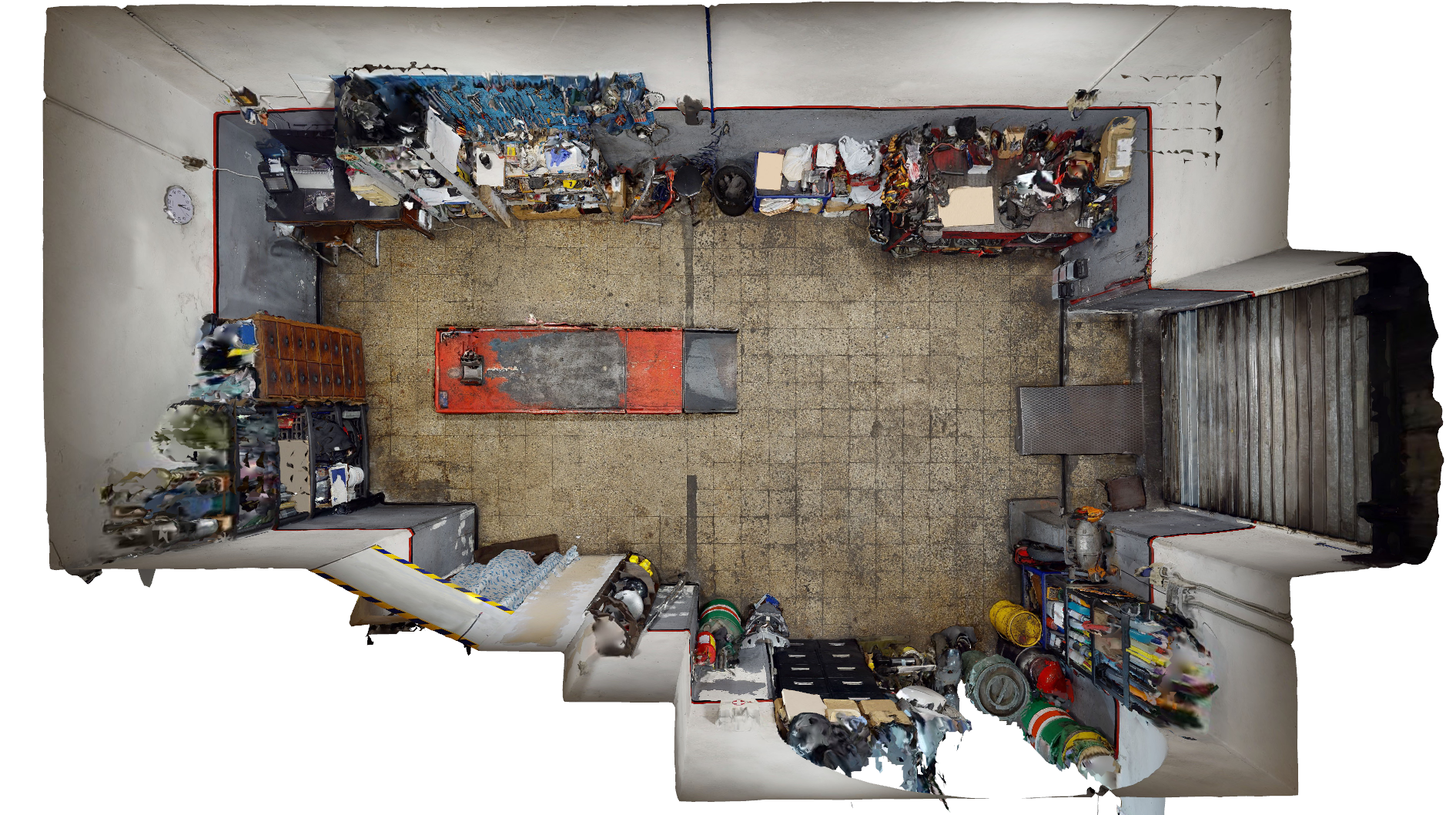}}
    \subfloat[Baker]{\includegraphics[width=0.4\textwidth]{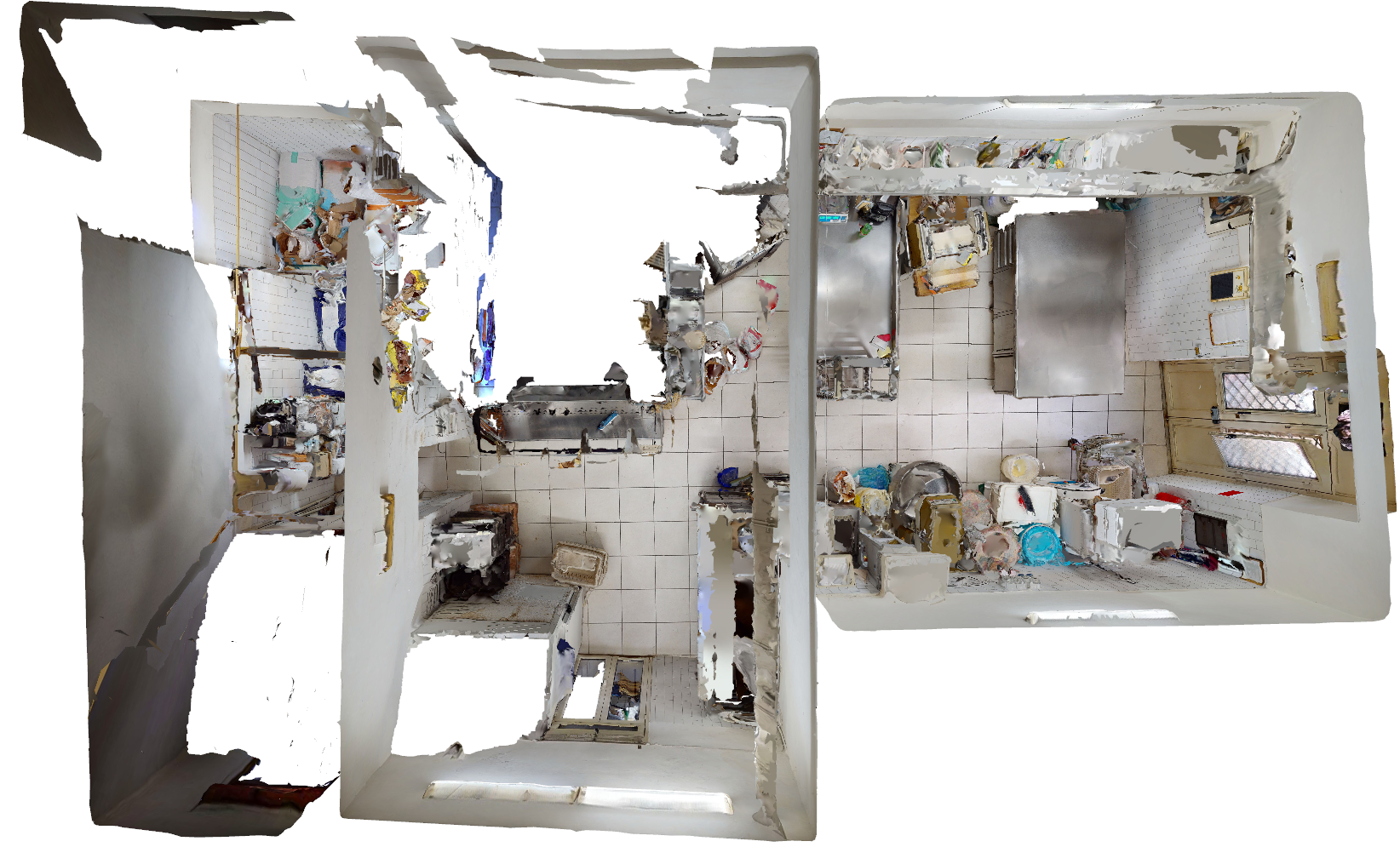}}
    
    \subfloat[Carpenter ]{\includegraphics[width=0.4\textwidth]{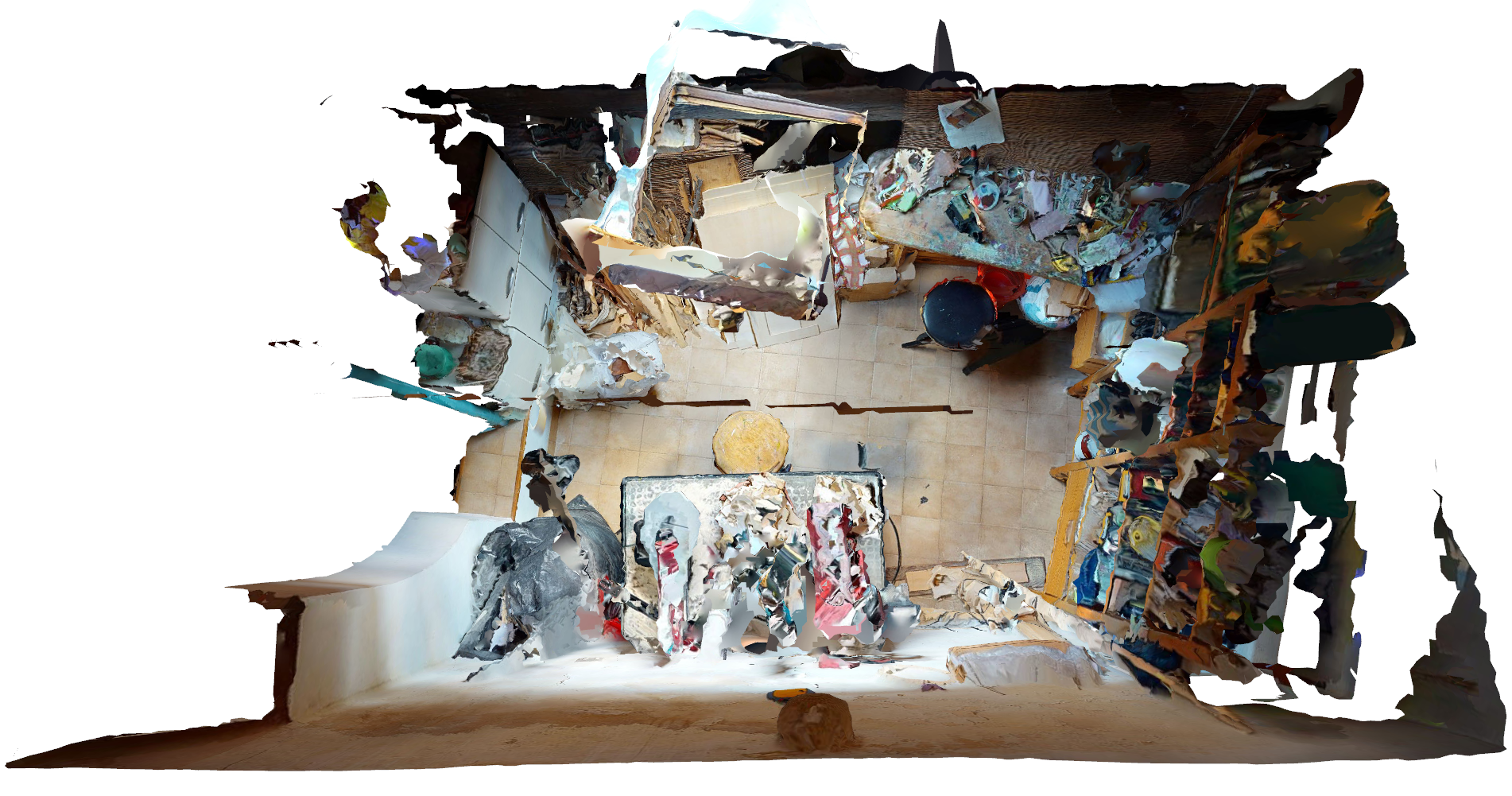}}
    \subfloat[Bike mechanic]{\includegraphics[width=0.4\textwidth]{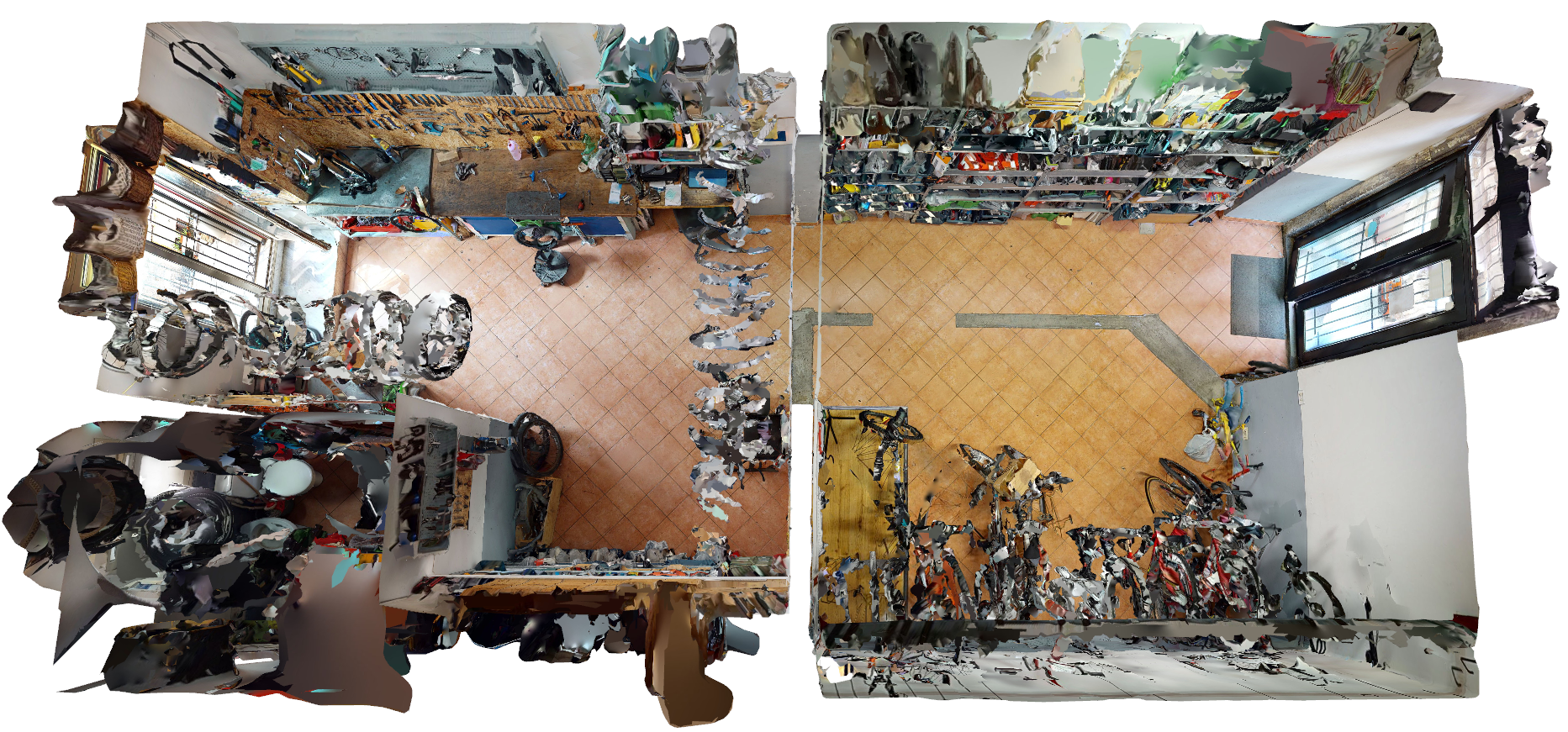}}
    
    \caption{\textbf{Top view for four examples of scene layout.}}
    \label{fig:four-scenes}
\end{figure*}

For a better illustration of the scene variance we have discussed in Sec.~\ref{sec:scene}, we show four examples of scene layout in Figure~\ref{fig:four-scenes}.

\end{document}